\documentclass[10pt,twocolumn,letterpaper]{article}

\usepackage[pagenumbers]{cvpr} % To force page numbers, e.g. for an arXiv version

\usepackage{graphicx}
\usepackage{amsmath}
\usepackage{amssymb}
\usepackage{booktabs}
\usepackage{multirow}
\usepackage{color}
\usepackage[dvipsnames]{xcolor}
\graphicspath{{./fig/}}
\usepackage{algorithmic}
\usepackage[ruled,vlined]{algorithm2e}
\usepackage{bbm}
\usepackage[pagebackref,breaklinks,colorlinks]{hyperref}

\usepackage[capitalize]{cleveref}
\crefname{section}{Sec.}{Secs.}
\Crefname{section}{Section}{Sections}
\Crefname{table}{Table}{Tables}
\crefname{table}{Tab.}{Tabs.}

\def\cvprPaperID{1561} % *** Enter the CVPR Paper ID here
\def\confName{CVPR}
\def\confYear{2022}

\newcommand{\myPara}[1]{\vspace{.05in}\noindent\textbf{#1}}

\newcommand{\bm}[1]{\mbox{\boldmath{$#1$}}}

\begin{document}

\title{Selective-Supervised Contrastive Learning with Noisy Labels}

%\author{
%	Shikun Li\\
%	%School of Cyber Security, University \\
%	%of Chinese Academy of Sciences\\
%	Institute of Information Engineering,\\
%	Chinese Academy of Sciences\\
%	{\tt\small lishikun@iie.ac.cn}\\  \\
%	Shiming Ge\\
%	%School of Cyber Security, University \\
%	%of Chinese Academy of Sciences\\
%	Institute of Information Engineering,\\
%	Chinese Academy of Sciences\\
%	{\tt\small geshiming@iie.ac.cn}
%	\and
%Xiaobo Xia\\
%The University of Sydney\\
%{\tt\small xiaoboxia.uni@gmail.com}\\
%\\ \\
%	Tongliang Liu\\
%	The University of Sydney\\
%	{\tt\small tongliang.liu@sydney.edu.au}
%}

\author{
Shikun Li$^{1,2}$, Xiaobo Xia$^{3}$, Shiming Ge$^{1,2}$\thanks{Shiming Ge is the corresponding author.}, Tongliang Liu$^{3}$ \\ 
$^{1}$ Institute of Information Engineering, Chinese Academy of Sciences, China \\
$^{2}$ School of Cyber Security, University of Chinese Academy of Sciences, China \\ 
$^{3}$ Trustworthy Machine Learning Lab, The University of Sydney, Australia \\
{\tt\small \{lishikun,geshiming\}@iie.ac.cn, xxia5420@uni.sydney.edu.au, tongliang.liu@sydney.edu.au}
}
\maketitle

%%%%%%%%% ABSTRACT
\begin{abstract}
Deep networks have strong capacities of embedding data into latent representations and finishing following tasks. However, the capacities largely come from high-quality annotated labels, which are expensive to collect. Noisy labels are more affordable, but result in corrupted representations, leading to poor generalization performance. To learn robust representations and handle noisy labels, we propose selective-supervised  contrastive learning (Sel-CL) in this paper. Specifically, Sel-CL extend supervised contrastive learning (Sup-CL), which is powerful in representation learning, but is degraded when there are noisy labels. Sel-CL tackles the direct cause of the problem of Sup-CL. That is, as Sup-CL works in a \textit{pair-wise} manner, noisy pairs built by noisy labels mislead representation learning. To alleviate the issue, we select confident pairs out of noisy ones for Sup-CL without knowing noise rates. In the selection process, by measuring the agreement between learned representations and given labels, we first identify confident examples that are exploited to build confident pairs. Then, the representation similarity distribution in the built confident pairs is exploited to identify more confident pairs out of noisy pairs. All obtained confident pairs are finally used for Sup-CL to enhance representations. Experiments on multiple noisy datasets demonstrate the robustness of the learned representations by our method, following the state-of-the-art performance. Source codes are available at \textcolor{magenta}{https://github.com/ShikunLi/Sel-CL}
\end{abstract}

%%%%%%%%% BODY TEXT
\section{Introduction}
Deep networks are powerful in various tasks, \textit{e.g.}, image recognition~\cite{he2016deep,Zhang2021TCSVT}, object detection~\cite{yang2022objects}, visual tracking~\cite{GeZL0T21} and text matching~\cite{ChenHNLLWX18}. The power is largely attributed to the collection of large-scale datasets with \textit{high-quality} annotated labels. In supervised learning, with the data (\textit{i.e.}, the instance and label pairs) in such datasets, deep networks first learn \textit{ideal latent representations} of the instances and then complete following tasks with the representations~\cite{goodfellow2016deep,yang2021free}. However, it is extremely expensive to obtain large-scale high-quality annotated labels. Alternatively, we can collect labels based
on web search and user tags~\cite{xiao2015learning,Li2017iccv}. These labels are cheap but inevitably noisy. 
\begin{figure}[!t]
    \centering
    \includegraphics[width=8.4cm,height=3.3cm]{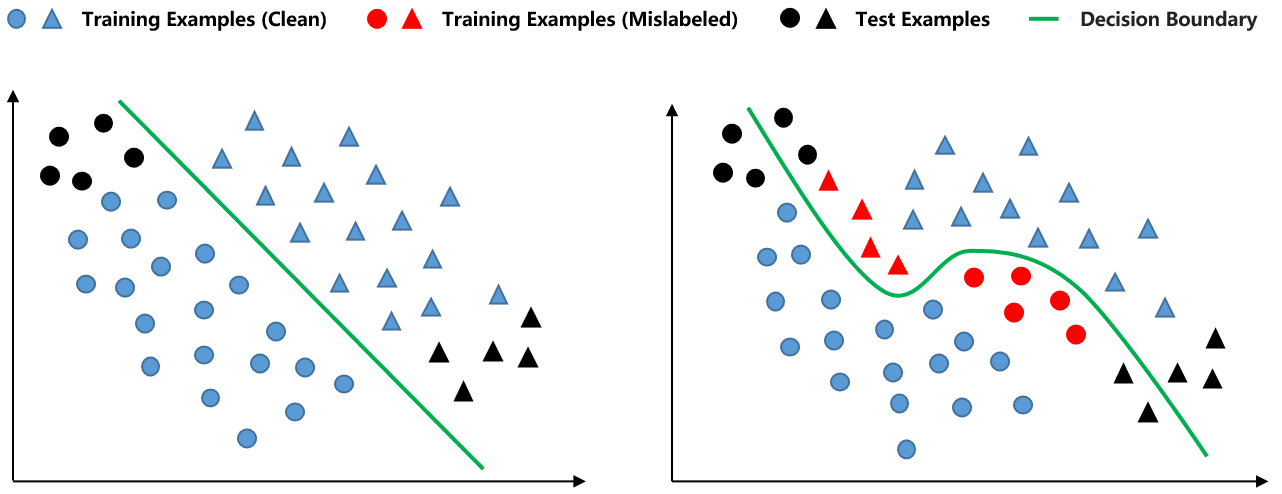}
    \caption{\textbf{Left}: learning a classifier with ideal representations induced by clean labels; \textbf{Right}: learning a classifier with corrupted representations caused by noisy labels. Circles represent the  representations of positive examples while triangles represent the representations of negative examples. When the representations are corrupted by noisy labels, the decision boundary of the classifier will be largely changed. Therefore, the learned classifier in this case cannot generalize well on test examples.}
    \vspace{-12pt}
    \label{fig:motivation}
\end{figure}

Noisy labels impair the generalization performance of deep networks~\cite{Zhang2017iclr,han2020sigua}. It is because, supervised by the datasets with noisy labels, the mislabeled data provide \textit{incorrect signals} when inducing latent representations for the instances. The \textit{corrupted representations} then cause inaccurate decisions for following tasks and hurt generalization \cite{lee2019robust,xia2021robust}. For example, as shown in Fig.~\ref{fig:motivation}, the \textit{corrupted representations} result in an inprecise classification boundary. Therefore, it is crucial to induce robust latent  representations of instances for learning with noisy labels, which is also \textit{our focus} in this paper.  

Recent works~\cite{chen2020big,chen2020simple,He0WXG20,Ghosh2021,Zheltonozhskii2021} show that, working in a \textit{pair-wise} manner, \textit{contrastive learning} (CL) methods can bring good latent representations to help following tasks. Based on whether \textit{supervised information} is provided, the CL methods can be grouped into supervised contrastive learning (Sup-CL)~\cite{Khosla2020} and unsupervised contrastive learning (Uns-CL)~\cite{chen2020big,chen2020simple,He0WXG20}. It has been shown that Sup-CL can exploit the supervised information to learn better representations than Uns-CL, but relies on the quality of supervised information~\cite{Ortego2021}. If the supervised information is corrupted by noisy labels, built pairs by training examples are noisy, following corrupted representations learned by Sup-CL. Motivated by this phenomenon, prior methods use \textit{general-purpose techniques} in tackling noisy labels for robust representation learning with Sup-CL, \textit{e.g.}, introducing regularization~\cite{Ortego2021} or generating pseudo-labels~\cite{Li2021ICLR}. Although these methods can work fine in some cases, the general-purpose techniques fail to consider the remarkable pair-wise characteristic of Sup-CL in strengthening representation learning. The achieved performance by them is thus argued to be \textit{sub-optimal}. 

In this paper, we propose \textit{selective-supervised contrastive learning} (Sel-CL) to address the above issue. Sel-CL can make use of the pair-wise characteristic to learn robust latent representations. The core idea of Sel-CL is (1) select confident pairs out of noisy pairs; (2) employ the confident pairs to learn robust latent representations. Note that it is hard to identify confident pairs directly for representation learning. The main reason is that we always need to set a threshold with the noise rate for precise identification, \textit{e.g.}, see \cite{Jiang2018icml,Han2018NIPS,han2020sigua}. Nevertheless, it is difficult to estimate the noise rate of noisy pairs. To handle this problem, we propose to first employ confident examples~\cite{Ortego2021,Han2018NIPS}, which is much easier to identified, to build a reliable set of confident pairs \textcolor{black}{at each epoch}. Then, based on the representation similarity distribution of confident pairs in this set, we set a dynamic threshold to selected more confident pairs out of all noisy pairs. By this pair-wise selection, we can make better use of not only the pairs whose class labels are correct, but also the pairs whose class labels are incorrect, but the examples in them are misclassified to the same class. All selected confident pairs are utilized to enhance representation learning with Sup-CL. 
As the selected confident pairs are less noisy, the learned representations \textcolor{black}{with this selective-supervised paradigm} will be more robust, naturally following promising generalization.
%the learned representations with them by Sup-CL will be more robust, naturally following promising generalization.

%We conduct a series of experiments on both simulated and real-world noisy datasets. The experimental results demonstrate that Sel-CL achieves state-of-the-art performance. Specifically, on simulated noisy datasets, within board ranges of noise rates and noise types, Sel-CL performs best in the vast majority of cases. On real-world noisy datasets, Sel-CL consistently performs best and is clearly ahead of existing advanced methods~\cite{LiSH20,Li2021ICLR,LiuNRF20}. Comprehensive ablation studies and discussions are also provided. 

The main contributions of this paper are summarized as three aspects:
1) We propose selective-supervised contrastive learning with noisy labels, which can obtain robust pre-trained representations by effectively selecting confident pairs for performing Sup-CL.
2) Without knowing the noise rate of pairs, our approach selects the pairs built by identified confident examples, and the pairs built by the examples with high representation similarities. It fulfils a positive cycle, where better confident pairs result in better representations and better representations will identify better confident pairs. 
3) We conduct experiments on synthetic and real-world noisy datasets, which clearly demonstrate our approach achieves better performance compared with the state-of-the-art methods. Comprehensive ablation studies and discussions are also provided. 

%The rest of this paper is organized as follows. In Section \ref{sec:2}, we briefly review related work with this paper. In Section \ref{sec:3}, we present our method step by step. In Section \ref{sec:4}, we provide comprehensive experimental evaluations to justify our claims. In Section \ref{sec:5}, we conclude this paper. 

\begin{figure*}
	\centering
	\includegraphics[width=1.0\linewidth,trim=5 5 5 5,clip]{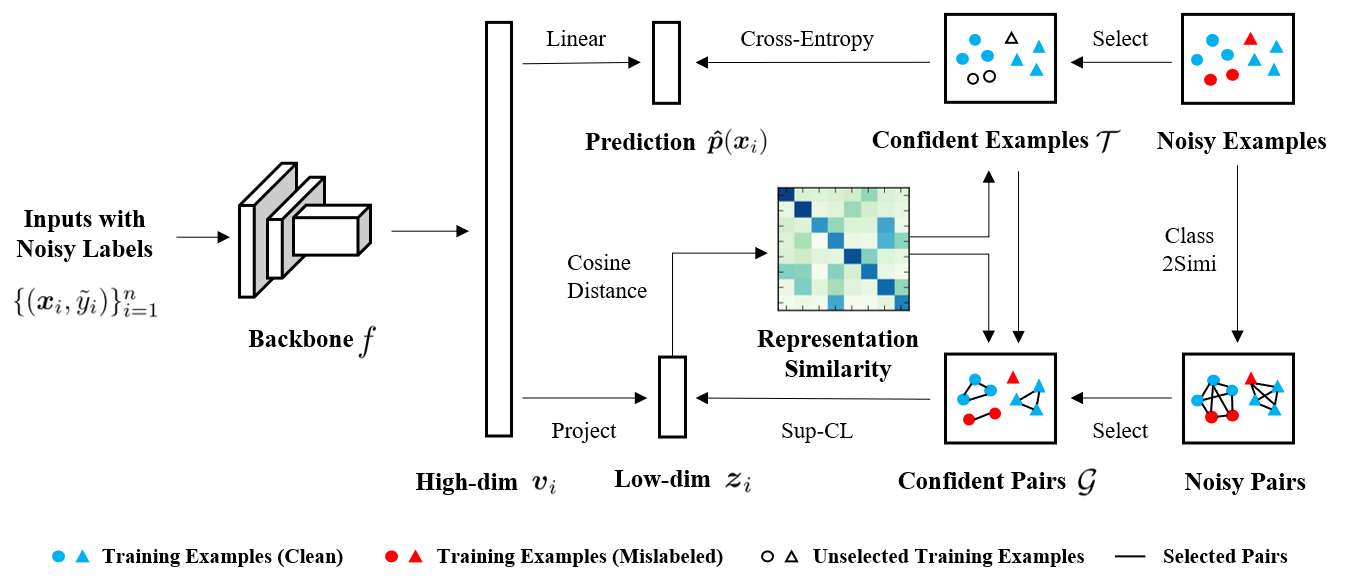}
	\caption{\textcolor{black}{The illustration of the proposed Sel-CL, which progressively selects better confident pairs $\mathcal{G}$ for supervised contrastive learning based on the representation similarity. Without the noise rate prior, confident examples $\mathcal{T}$ are also obtained to help identify the pairs.}}
	\label{fig:pipeline}
\end{figure*}

\section{Related Works}\label{sec:2}
In this section, we review recent methods on learning with noisy labels and contrastive learning. 

\subsection{Learning with Noisy Labels}
There are a large body of recent works on learning with noisy labels, which include but do not limit to estimating the noise transition
matrix~\cite{HendrycksMWG18,xia2019anchor,xia2020part,cheng2022cvpr}, reweighting examples~\cite{Liu2016TPAMI,RenZYU18,ShuXY0ZXM19,wang2021graph}, selecting confident examples~\cite{huang2019o2u,YangICML2020,chen2022anomman,Li2020aaai}, designing robust loss functions~\cite{zhang2018generalized,GhoshKS17,ChengZLGSL21,wei2021robust}, introducing regularization~\cite{zhang2018mixup,hu2020simple,chen2021noise}, generating pseudo labels~\cite{tanaka2018joint,zheng2020error,zhang2021learningwith,han2019deep,Li2021TMM}, and \textit{etc}. In addition, some advanced start-of-the-art methods combine serveral techniques, \textit{e.g.}, DivideMix~\cite{LiSH20} and ELR+~\cite{LiuNRF20}.

\subsection{Contrastive Learning}
Recent works in unsupervised contrastive learning (Uns-CL)~\cite{chen2020simple,chen2020big,He0WXG20,wang2021cris} have demonstrated the potential of contrastive based similarity learning frameworks for representation learning. These methods maximize (minimize) similarities of positive (negative) pairs at the \textit{instance level}. That is to say, the positive pair is only built by two correlated views of the same instance. The other data pairs are negative. To make use of supervised information for learning better representations, Uns-CL is extended to the fully-supervised setting, named supervised contrastive learning (Sup-CL) ~\cite{Khosla2020}. Sup-CL aims to make examples belonging to the same class lie closer in the representation space than those of examples from different classes. With clean labels, Sup-CL achieves promising performance. 

In consideration of the strong power of Sup-CL, some works tend to use it to learn latent representations and handle noisy labels. For example, the methods MOIT+~\cite{Ortego2021} and MoPro~\cite{Li2021ICLR} work in two stages. First, they pre-train the networks with Sup-CL and use general-purpose techniques, \textit{e.g.}, adding regularization, to reduce the side effect of noisy labels. Second, the network is fine-tuned on a reliable dataset. Additionally, the methods ProtoMix~\cite{li2020learning} and NGC~\cite{Zhi2021ICCV} work in one stage. They jointly perform the generation of pseudo labels and Sup-CL to combat noisy labels. In this paper, we follow the two-stage learning style. The \textit{pair-wise} characteristic of contrastive learning is investigated to enhance representation learning and further better handle noisy labels. 
\vspace{-3pt}
\section{Selective-Supervised Contrastive Learning}\label{sec:3}
%\vspace{-3pt}
%In this section, we first introduce the used notations and problem setting and then present our method step by step. 
%
%\subsection{Preliminaries}
%We begin by fixing notations. Scalars are in lowercase letters. Vectors are in lowercase boldface letters. Let $\mathbbm{I}[A]$ be the indicator of the event $A$. Let $[z]=\{1,\ldots,z\}$. 
%
%Consider a classification task, there are $C$ classes. We are given a noisily-labeled dataset $\widetilde{\mathcal{D}}= \{ (\bm{x}_i,\tilde{y}_i )\}_{i=1}^n$, where $n$ is the sample size, $\bm{x}_i$ is the instance of the $i$-th example and $\tilde{y}_i \in [C]$ is the corresponding noisy label. For $\tilde{y}_i$, the associated but unobservable true label is denoted by $y_i$. The aim is to train a deep network robustly with a noisy train dataset and obtain high accuracy on a clean test dataset. 
%
%The proposed method works in two stages. During pre-training stage, the used network consists of three components: (1) a deep encoder $f$ (a convolutional neural network backnone) that the instance $\bm{x}_i$ to a high-dimensional representation $\bm{v}_i$; (2) a classifier head (a fully-connected layer followed by the softmax function) that receives $\bm{v}_i$ as an input and outputs class predictions $\bm{\hat{p}}(\bm{x}_i)$; (3) a linear or non-linear projection that maps  $\bm{v}_i$ into a low-dimensional representation $\bm{z}_i$. During the fine-tuning stage, we only keep the pre-trained deep encoder, and apply one new classifier head on the top to output the predictions. In the following, we present our method. 

We begin by fixing notations. Scalars are in lowercase letters. Vectors are in lowercase boldface letters. Let $\mathbbm{I}[A]$ be the indicator of the event $A$. Let $[z]=\{1,\ldots,z\}$. 
Consider a classification task, there are $C$ classes. We are given a noisily-labeled dataset $\widetilde{\mathcal{D}}= \{ (\bm{x}_i,\tilde{y}_i )\}_{i=1}^n$, where $n$ is the sample size, $\bm{x}_i$ is the instance of the $i$-th example and $\tilde{y}_i \in [C]$ is the corresponding noisy label. For $\tilde{y}_i$, the associated but unobservable true label is denoted by $y_i$. 

\myPara{Overview.} Following the two-stage learning style~\cite{Ortego2021,Li2021ICLR,Ghosh2021,Zheltonozhskii2021}, our aim is to induce robust pre-trained representations of instances for learning with noisy labels.

\textcolor{black}{To achieve it, our approach progressively selects better confident pairs $\mathcal{G}$ out of noisy pairs for performing supervised contrastive learning at each epoch. Without the noise rate prior, to help identify the pairs, confident examples $\mathcal{T}$ are also obtained in this process. The used pre-training network consists of three components: (1) a deep encoder $f$ (a convolutional neural network backnone) that maps the instance $\bm{x}_i$ to a high-dimensional representation $\bm{v}_i$; (2) a classifier head (a fully-connected layer followed by the softmax function) that receives $\bm{v}_i$ as an input and outputs class predictions $\bm{\hat{p}}(\bm{x}_i)$; (3) a linear or non-linear projection that maps  $\bm{v}_i$ into a low-dimensional representation $\bm{z}_i$.  The illustration of the proposed Sel-CL can be seen in Fig.~\ref{fig:pipeline}.
After that, with the robust representations, we only keep the pre-trained deep encoder, and apply one new classifier head on the top to output the predictions for fine-tuning stage. In the following, we present our method step by step.}

%The aim is to train a deep network robustly with a noisy train dataset and obtain high accuracy on a clean test dataset. 
%
%The proposed method works in two stages. During pre-training stage, the used network consists of three components: (1) a deep encoder $f$ (a convolutional neural network backnone) that the instance $\bm{x}_i$ to a high-dimensional representation $\bm{v}_i$; (2) a classifier head (a fully-connected layer followed by the softmax function) that receives $\bm{v}_i$ as an input and outputs class predictions $\bm{\hat{p}}(\bm{x}_i)$; (3) a linear or non-linear projection that maps  $\bm{v}_i$ into a low-dimensional representation $\bm{z}_i$. During the fine-tuning stage, we only keep the pre-trained deep encoder, and apply one new classifier head on the top to output the predictions. In the following, we present our method.
\vspace{-3pt}
\subsection{Selecting Confident Examples}
\textcolor{black}{To help identify the confident pairs, we first select confident examples base on the representation similarity.
In addition, we warm up the training in the first few epochs to obtain  low-dimensional representations of instances for identifying confident examples later. Specifically, we use Uns-CL \cite{chen2020simple} for the warm-up training. Note that our method is robust to the choice of warm-up methods (See Section~\ref{sec:4.6}).} %The details will be shown in experiments. 

The confident examples are identified by measuring agreements between the obtained low-dimensional representations and given labels. For this goal, given two low-dimensional representations $\bm{z}_i$ and $\bm{z}_j$, we first calculate the representation similarity between them by the cosine distance, \textit{i.e.},
\begin{equation}\label{eq:re_sim}
	d\left(\bm{z}_{i}, \bm{z}_{j}\right)=\frac{\bm{z}_{i}\bm{z}_{j}^{\top}}{\left\|\bm{z}_{i}\right\|\left\|\bm{z}_{j}\right\|}.
\end{equation}
Then, to quantify the agreement, as did in \cite{Ortego2021}, for each example $(\bm{x}_i,\tilde{y}_i)$, we aggregate the original label from its top-$K$ neighbors based on the representation similarity to create a pseudo-label $\hat{y}_i$. That is to say, we count the original labels of its top-$K$ neighbors ($K$=250 in the experiements) and correct $\tilde{y}_i$ using the dominant class. In this way, we can better use of representation similarities to improve the detection of mislabeled examples~\cite{Ortego2021}. We use the pseudo-labels to approximate clean class posterior probabilities, \textit{i.e.}, %Since the pseudo-labels is less noisy, 
\begin{equation}
	\hat{q}_c\left(\bm{x}_{i}\right)=\frac{1}{K} \sum_{k=1 \atop \bm{x}_{k} \in \mathcal{N}_{i}}^{K} \mathbbm{I}[\hat{y}_{k} = c], c\in[C],
\end{equation}	
where $\mathcal{N}_{i}$ denotes the neighbor set of $K$ closest instances to $\bm{x}_i$ according to the learned representation. Following \cite{Han2018NIPS}, we exploit the cross-entropy loss $\ell$ to identify confident examples. Denoted the set of confident examples belonging to the $c$-th class as $\mathcal{T}_c$, we have 
\begin{equation}
	\mathcal{T}_c=\{(\bm{x}_i,\tilde{y}_i)\mid \ell(\bm{\hat{q}}(\bm{x}_i), \tilde{y}_i)\textless \gamma_c, i \in [n] \}, c\in[C],
	\label{class_label}
\end{equation}	
where $\gamma_c$ is a threshold for the $c$-th class, which is dynamically defined to ensure a class-balanced set of identified confident examples. To achieve this goal, we use the $\alpha$ fractile of per-class agreements between the corrected label $\hat{y}_i$ and the original label $\tilde{y}_i$ across all classes to determine how many examples should be selected for each class, i.e., $\sum_{i=1}^n \mathbb{I}[\hat{y}_{i}=\tilde{y}_{i}][\tilde{y}_{i}=c], c \in[C]$. Finally, we can get the confident example set including all classes, \textit{i.e.}, $\mathcal{T}= \cup_{c=1}^{C} \mathcal{T}_c$. This set is less noisy than original noisy datasets and therefore more reliable.

\subsection{Selecting Confident Pairs}
As it is hard to achieve a precise estimation of the noise rate of noisy pairs, we use identified confident examples to help select confident pairs. Specifically, we transform identified confident examples into a set of associated confident pairs.  Denoted this set by $\mathcal{G}^{\prime}$, we have 
\begin{equation}
	\mathcal{G}^{\prime}=\{P_{ij}\mid \tilde{y}_{i} = \tilde{y}_{j}, (\bm{x}_i,\tilde{y}_i), (\bm{x}_j,\tilde{y}_j) \in \mathcal{T} \}, 
	\label{simi_label1}
\end{equation}
where $P_{ij}$ is the pair built by the examples $(\bm{x}_i,\tilde{y}_i)$ and $(\bm{x}_j,\tilde{y}_j)$. As $\mathcal{G}^{\prime}$ is built by $\mathcal{T}$, it is reliable. It should be noted that, even though two examples in one pair have incorrect class labels, the similarity labels are still correct when two examples are misclassified to the same class \cite{Wu2021icml}. Such pairs are valuable for representation learning. We thus identify more confident pairs from noisy pairs. We define whether two instances $\bm{x}_i$ and $\bm{x}_j$ belong to the same class as their similarity label, \textit{i.e.}, $\tilde{s}_{ij}=\mathbbm{I}[\tilde{y}_{i} = \tilde{y}_{j}]$. Note that similarity labels are extremely imbalanced, \textit{i.e.}, the positive ones are less than negative ones \cite{Wu2021icml}. For convenience, we consider the noisy pairs with positive similarity labels. Denoted by the set of confident pairs identified from noisy positive pairs by $\mathcal{G}^{\prime\prime}$. We have 
\begin{equation}
	\mathcal{G}^{\prime\prime}=\{P_{ij}\mid \tilde{s}_{ij}=1, d\left(\bm{z}_{i}, \bm{z}_{j}\right)>\gamma \},
	\label{simi_label2}
\end{equation}
where $\gamma$ is a dynamic threshold to control the number of identified confident pairs; $i$ and $j$ are two indices sampled from all training data. To avoid the noise rate estimation of noisy positive pairs, we utilize the reliable information of $\mathcal{T}$ to set $\gamma$. In more detail, the $\beta$ fractile of the representation similarities of the pairs in $\mathcal{G}^{\prime}$ is used here. Finally, we can get the confident pair set $\mathcal{G}= \mathcal{G}^{\prime} \cup \mathcal{G}^{\prime\prime}$. 
\subsection{Representation Learning with Selected Pairs}
After selecting confident pairs, we can learn representations by utilizing them with supervised contrastive learning at each epoch. As selected confident pairs can be less noisy, this selective-supervised paradigm can enhance representations to handle noisy labels.

\myPara{Contrastive learning.} Following the original Sup-CL \cite{Khosla2020}, we randomly sample $N$ instances in each mini-batch to apply two random data augmentation operations to each instance, thus generating two data views. The resulting training mini-batch data is $\{(\bm{x}_i, \tilde{y}_i)\}^{2N}_{i=1}$, where $i \in I=[2N]$ is the index of an arbitrary augmented instance.
Given $\mathcal{G}$, we perform supervised contrastive learning with selected confident positive pairs:
\begin{equation}\label{eq6}
	\begin{split}
	\mathcal{L}=&\sum_{i \in I}\mathcal{L}_i(\bm{z}_i)\\
	=&\sum_{i \in I} \frac{-1}{|\mathcal{G}(i)|} \sum_{g \in \mathcal{G}(i)} \log \frac{\exp \left(\bm{z}_i \cdot \bm{z}_g / \tau\right)}{\sum_{a \in A(i)} \exp \left(\bm{z}_i \cdot \bm{z}_a / \tau\right)},
\end{split}
\end{equation}
where $A(i)$ means the set of indices excluding $i$, i.e., $A(i) = I \backslash\{i\}$; $\mathcal{G}(i) = \left\{g \mid g\in A(i), P_{i^{\prime}g^{\prime}} \in \mathcal{G} \right\}$, and $i^{\prime}$ and $g^{\prime}$ are the original indices of $\bm{x}_i$ and $\bm{x}_g$ in $\widetilde{\mathcal{D}}$, respectively. $\tau \in \mathbbm{R}^{+}$ is a temperature parameter. Note that expect for the examples that are involved in selected confident pairs, for the other examples, we perform unsupervised contrastive learning \cite{chen2020simple} on them.

Additionally, to further make representation learning robust, following MOIT~\cite{Ortego2021}, a Mixup technique~\cite{zhang2018mixup} is added into our framework. Mixup performs convex combination of pairs of examples as $\bm{x}_{i}=\lambda \bm{x}_{a}+(1-\lambda) \bm{x}_{b}$, where 
$\lambda \in[0,1] \sim \operatorname{Beta}(\alpha_m, \alpha_m)$ and $\bm{x}_i$ denotes the training example that combines two mini-batch examples $\bm{x}_a$ and $\bm{x}_b$. A linear relation in the contrastive loss is imposed as 
\begin{equation}
	\mathcal{L}_{i}^{\mathrm{MIX}}(\bm{z}_i)=\lambda \mathcal{L}_{a}(\bm{z}_i)+(1-\lambda) \mathcal{L}_{b}(\bm{z}_i),
\end{equation}
where  $\mathcal{L}_{a}$ and $\mathcal{L}_{b}$ have the same
form as $\mathcal{L}_{i}$ in Eq.~\ref{eq6}. It should be noted that the selection of positive/negative
pairs involves considering a unique label for each mixed
example \cite{Ortego2021}. Nevertheless, the input example always contain
two labels, where $\lambda$ determines the dominant one. We assign this dominant label to every example for positive/negative sampling.

\myPara{Classification learning.}
A classification objective with confident examples is also employed to stabilize the convergence and achieve better representations. Given the confident examples of $\mathcal{T}$, classification learning is conducted by using
\begin{equation}
	\mathcal{L}^{\mathrm{CLS}}=\sum_{(\bm{x}_i,\tilde{y}_i) \in \mathcal{T}}\mathcal{L}^{cls}_i(\bm{x}_i)=\sum_{(\bm{x}_i,\tilde{y}_i) \in \mathcal{T}}\ell(\bm{\hat{p}}(\bm{x}_i), \tilde{y}_i),
\end{equation}
where $\bm{x}_i$ can also refer to the augmented image.

Moreover, inspired by recent methods~\cite{Wu2021icml,Hsu2019iclr} that learn classifiers with similarity labels, we add a learning objective to directly learn from similarity labels using classfier predictions. Given a multiviewed mini-batch data $\{(\bm{x}_i, \tilde{y}_i)\}^{2N}_{i=1}$, the added similarity loss is: 
\begin{equation}
	\mathcal{L}^{\mathrm{SIM}}= \sum_{i \in I}\sum_{j \in A(i)} \ell(\bm{\hat{p}}(\bm{x}_i)\bm{\hat{p}}(\bm{x}_j),\mathbbm{I}[P_{i^{\prime}j^{\prime}} \in \mathcal{G}]).
\end{equation}

Lastly, combining the above analyses, the total objective loss is: 
\begin{equation}
	\mathcal{L}^{\mathrm{ALL}}=\mathcal{L}^{\mathrm{MIX}}+\lambda_c\mathcal{L}^{\mathrm{CLS}}+\lambda_s \mathcal{L}^{\mathrm{SIM}},
	\label{all} 
\end{equation}
where $\lambda_c$ and $\lambda_s$ are loss weights, which we set as $\lambda_c=1,\lambda_s=0.01$ in all experiments. Note that, by alternately identifying confident pairs and learning robust representations at each epoch,  it fulfils a positive cycle that better confident pairs will result in better learned representations and better representations will identify better confident pairs. The similar idea of the positive cycle is shared by \cite{bai2020me}. %The illustration of the proposed Sel-CL can be seen in Fig.~\ref{fig:pipeline}.
%With such representation learning and noise detection performed alternately each epoch, it fulfils a positive cycle that better confident pairs will result in better learned representations and better representations will identify better confident pairs.

\subsection{Classification Fine-tuning}
With the pre-trained robust representation before, we only keep the pre-trained deep encoder $f$, and apply one new classifier head on top to form a classifer network $f'$ and output the predictions. The existing methods on handling noisy labels can be can be applied to fine-tune the classifier. To avoid the complexity of the method, we fine-tune the deep networks on identified confident examples, with a simple robust loss function~\cite{Ortego2021}. In order to better distinguish the two stages in our framework, we call the proposed method with fine-tuning as Sel-CL+. Also, we empirically verify that our method is also applicable with complex fine-tuning algorithms, \textit{e.g.}, DivideMix~\cite{LiSH20} and ELR+~\cite{LiuNRF20}. %The illustration of the proposed Sel-CL can be seen in Fig.~\ref{fig:pipeline}.
%We summarize the overall procedure of our method in Algorithm [ref]. 

\vspace{-5pt}
\section{Experiments}\label{sec:4}
In this section, we employ comprehensive experiments to verify the effectiveness of our method.
\subsection{Datasets and Implementation Details}\label{sec:4.1}
\vspace{-5pt}
\myPara{Simulated noisy datasets.}
We validate our method on two simulated noisy datasets, \textit{i.e.}, CIFAR-10~\cite{krizhevsky2009learning} and CIFAR-100~\cite{krizhevsky2009learning}. Both CIFAR-10 and CIFAR-100 contain 50k training images and 10k test images of the size $32\times32\times3$. Following previous works~\cite{tanaka2018joint, LiSH20,Ortego2021}, we consider two settings of simulated noisy labels: symmetric and asymmetric noise. Symmetric noise is generated by randomly replacing the labels for a percentage of the training data with all possible labels. Asymmetric noise uses label flips to incorrect classes: ``truck $\rightarrow$ automobile, bird $\rightarrow$ airplane, deer $\rightarrow$ horse, cat $\leftrightarrow$ dog'' in CIFAR-10, whereas in CIFAR-100 label flips are done circularly within the super-classes. 

For CIFAR-10/100 datasets , we use a PreAct ResNet-18 network, and train it using SGD with a momentum of 0.9, a weight decay of $10^{-4}$, and a batch size of 128. The network is trained for 250 epochs and the warm-up training has 1 epoch. We set the initial learning rate as 0.1, and reduce it by a factor of 10 after 125 and 200 epochs. The fine-tuning stage of Sel-CL+ has 70 epochs, where the learning rate is 0.001. We always use the Mixup parameter $\alpha_m=1$, scalar temperature $\tau=0.1$, and loss weights $\lambda_c=1,\lambda_s=0.01$. The noise detection fractiles on the synthetic datasets are shown in Tab.~\ref{fractiles}. We apply the strong data augmentations SimAug~\cite{chen2020simple} in the pre-training stage, and the standard weak data augmentations in the fine-tuning stage. To save computing resources, the MOCO trick~\cite{He0WXG20} is used, and the size of queue is set to 30k.
\begin{table}[t]		
	\caption{Noise detection fractiles on simulated noisy CIFAR-10 and CIFAR-100.}
	\centering
	\small
	\begin{tabular}{c|c|c|c|c}
		\hline \multirow{2}{*}{Fractiles} &\multicolumn{2}{c}{ CIFAR-10} & \multicolumn{2}{|c}{CIFAR-100} \\
		\cline{2-5}
		&  Sym. & Asym.  & Sym. & Asym. \\
		\hline
		$\alpha$ &  50\% & 50\% & 75\% & 25\% \\
		$\beta$ & 25\% & 25\% & 35\% & 0\% \\			
		\hline
	\end{tabular}
	\label{fractiles}
	\vspace{-3pt}
\end{table}

\begin{table}		
	\caption{Weighted KNN evaluations (\%) on CIFAR-100. The best results are in \textbf{bold}.}
	\centering
	\small
	\begin{tabular}{c|c|c|c|c|c}
		\hline \multirow{2}{*}{Methods} & Clean &\multicolumn{2}{c}{ Symmetric} & \multicolumn{2}{|c}{Asymmetric} \\\cline{2-6}
		& 0\% & 20\% & 80\% & 10\% & 40\% \\
		\hline
		Uns-CL~\cite{chen2020simple} & \underline{56.23} & -- & -- & -- & -- \\
		Sup-CL~\cite{Khosla2020} & 72.66 & 58.32 & 41.00 & 71.11 & 68.00 \\
		MOIT~\cite{Ortego2021} & \underline{77.48} & 67.42 & 55.58 & 74.86 & 72.60 \\
		Sel-CL & \textbf{77.94} & \textbf{75.36} & \textbf{62.49} & \textbf{76.77} & \textbf{72.71} \\				
		\hline
	\end{tabular}
	\label{knn}
	\vspace{-6pt}
\end{table}

\myPara{Real-world noisy dataset.} We validate our method on a real-world noisy dataset, \textit{i.e.}, WebVision~\cite{abs_1708_02862}. WebVision contains 2.4 million images crawled from the web using the 1,000 concepts in ImageNet ILSVRC12. Following previous works~\cite{LiSH20,Ortego2021}, we compare baseline methods on the first 50 classes of the Google image subset, called WebVision-50. For WebVision-50, we use a standard ResNet-18, and train it using SGD with a momentum of 0.9, a weight decay of $10^{-4}$, and a batch size of 64. The network is trained for 130 epochs and the warm-up training has 5 epochs. We set the initial learning rate as 0.1, and reduce it by a factor of 10 after 80 and 105 epochs. The fine-tuning stage has 50 epochs, where the learning rate is 0.001. We use the Mixup parameter $\alpha_m=1$, scalar temperature $\tau=0.1$, loss weights $\lambda_c=1,\lambda_s=0.01$. The noise detection fractiles are  $\alpha=40\%$ and $\beta=0\%$. The data augmentation techniques are similar to those of the CIFAR-10/100 datasets. The MOCO trick~\cite{He0WXG20} also is used, and the size of queue is 60k.

For all baselines, we use similar configurations for the same datasets for a fair comparison. We acquire the results of some baselines based on published codes. The results with \underline{underlines} mean that we obtain them based on published codes. Implementation details for baselines are provided in Appendix B. We also borrow other experimental results from the related works~\cite{LiSH20,Zheltonozhskii2021,LiuNRF20, Zhi2021ICCV,Ortego2021icpr,Ortego2021}.

\vspace{-2pt}
\subsection{Representation Learning Evaluations}\label{sec:4.2}

\begin{table*}[h]
	\caption{Comparison with state-of-the-art methods in the test accuracy (\%) on CIFAR-10 and CIFAR-100. The best results are in \textbf{bold}.} %{*} denotes results acquired by us based on published code.Results of other baselines are from~\cite{LiSH20,Ortego2021icpr,Ortego2021}.
	\centering
	\small
	\setlength\tabcolsep{3.5pt}
	\begin{tabular}{l|cccc|cccc|cccc|cccc}
		\hline \multirow{1}{*}{Dataset} & \multicolumn{8}{c}{ CIFAR-10} & \multicolumn{8}{|c}{CIFAR-100} \\
		\hline \multirow{2}{*}{Methods/Noise rate} & \multicolumn{4}{c}{ Symmetric} & \multicolumn{4}{|c|}{Asymmetric}  & \multicolumn{4}{c}{ Symmetric} & \multicolumn{4}{|c}{Asymmetric} \\\cline{2-17}
		& 20\% & 50\% & 80\% & 90\% & 10\% & 20\% & 30\% & 40\% & 20\% & 50\% & 80\% & 90\% & 10\% & 20\% & 30\% & 40\% \\
		\hline	
		Cross-Entropy & 82.7 & 57.9 & 26.1 & 16.8 & 88.8 & \underline{86.1} & 81.7 & 76.0 & 61.8 & 37.3 & 8.8 & 3.5 & 68.1 & \underline{63.6} & 53.3 & 44.5 \\	
		Mixup~\cite{zhang2018mixup} & 92.3 & 77.6 & 46.7 & 43.9 & 93.3 & \underline{88.0} & 83.3 & 77.7  & 66.0 & 46.6 & 17.6 & 8.1 & 72.4 & \underline{65.1} & 57.6 & 48.1\\			
		Forward~\cite{PatriniRMNQ17} & 83.1 & 59.4 & 26.2 & 18.8 & 90.4 & \underline{86.7} & 81.9 & 76.7  & 61.4 & 37.3 & 9.0 & 3.4 & 68.7 & \underline{63.2} & 54.4 & 45.3\\		
		\underline{GCE}~\cite{zhang2018generalized} & 86.6 & 81.9 & 54.6 & 21.2 & 89.5 & 85.6 & 80.6 & 76.0 & 59.2 & 47.8 & 15.8 & 7.2 & 68.0 & 58.6 & 51.4 & 42.9\\
		P-correction~\cite{YiW19} & 92.0 & 88.7 & 76.5 & 58.2 & 93.1 & \underline{92.9} & 92.6 & 91.6 & 68.1 & 56.4 & 20.7 & 8.8 & 76.1 & \underline{68.9} & 59.3 & 48.3\\ 		
		M-correction~\cite{ArazoOAOM19} & 93.8 & 91.9 & 86.6 & 68.7 & 89.6 & \underline{91.8} & 92.2 & 91.2  & 73.4 & 65.4 & 47.6 & 20.5 & 67.1 & \underline{64.5} & 58.6 & 47.4 \\             					
		DivideMix~\cite{LiSH20} & 95.0 & {93.7} & \textbf{92.4} & 74.2 & 93.8 & \underline{93.2} & 92.5 & 91.4 & 74.8 & {72.1} & 57.6 & 29.2 & 69.5 & \underline{69.2} & 68.3 & 51.0 \\		
		\underline{ELR}~\cite{LiuNRF20} & 93.8 & 92.6 & 88.0 & 63.3 &  94.4 & 93.3 & 91.5 & 85.3 & 74.5 & 70.2 & 45.2 & 20.5 & 75.8 & 74.8 & 73.6 & 70.0 \\
		\hline		
		\underline{GCE (Uns-CL init.)}~\cite{Ghosh2021} & 90.0 & 89.3 & 73.9 & 36.5 & 91.1 & 87.3 & 82.2 & 78.1 & 68.1 & 53.3 & 22.1 & 8.9 & 70.2 & 60.2 & 52.6 & 44.1 \\		
		\underline{ELR (Uns-CL init.)} & 94.4 & 93.0 & 88.3 & \textbf{86.2} &  {95.0} & {94.7} & {94.4} &  {93.3} & {76.2} & 71.9 & {57.9} & 40.8 & 77.2 & 75.5 & 74.3 & 70.4\\
		MOIT+~\cite{Ortego2021} & \underline{94.1} & \underline{91.8} & \underline{81.1} & \underline{74.7} & 94.2 & \underline{94.3} & 94.3 & 93.3 & {75.9} & \underline{70.6} & {47.6} & \underline{41.8} & 77.4 & \underline{76.4} & 75.1 & 74.0 \\
		\hline			
		Sel-CL+ & \textbf{95.5} & \textbf{93.9} & 89.2 & 81.9 &  \textbf{95.6} &  \textbf{95.2} & \textbf{94.5}  & \textbf{93.4} & \textbf{76.5} & \textbf{72.4} & \textbf{59.6} & \textbf{48.8} & \textbf{78.7} &  \textbf{77.5} & \textbf{76.4}  & \textbf{74.2} \\
		\hline
	\end{tabular}
	\label{cifar10}
	%\vspace{-5pt}
\end{table*}

\begin{figure}[h]
	\centering
	\subfloat[]{\includegraphics[width=1.65in,trim=30 0 0 0,clip]{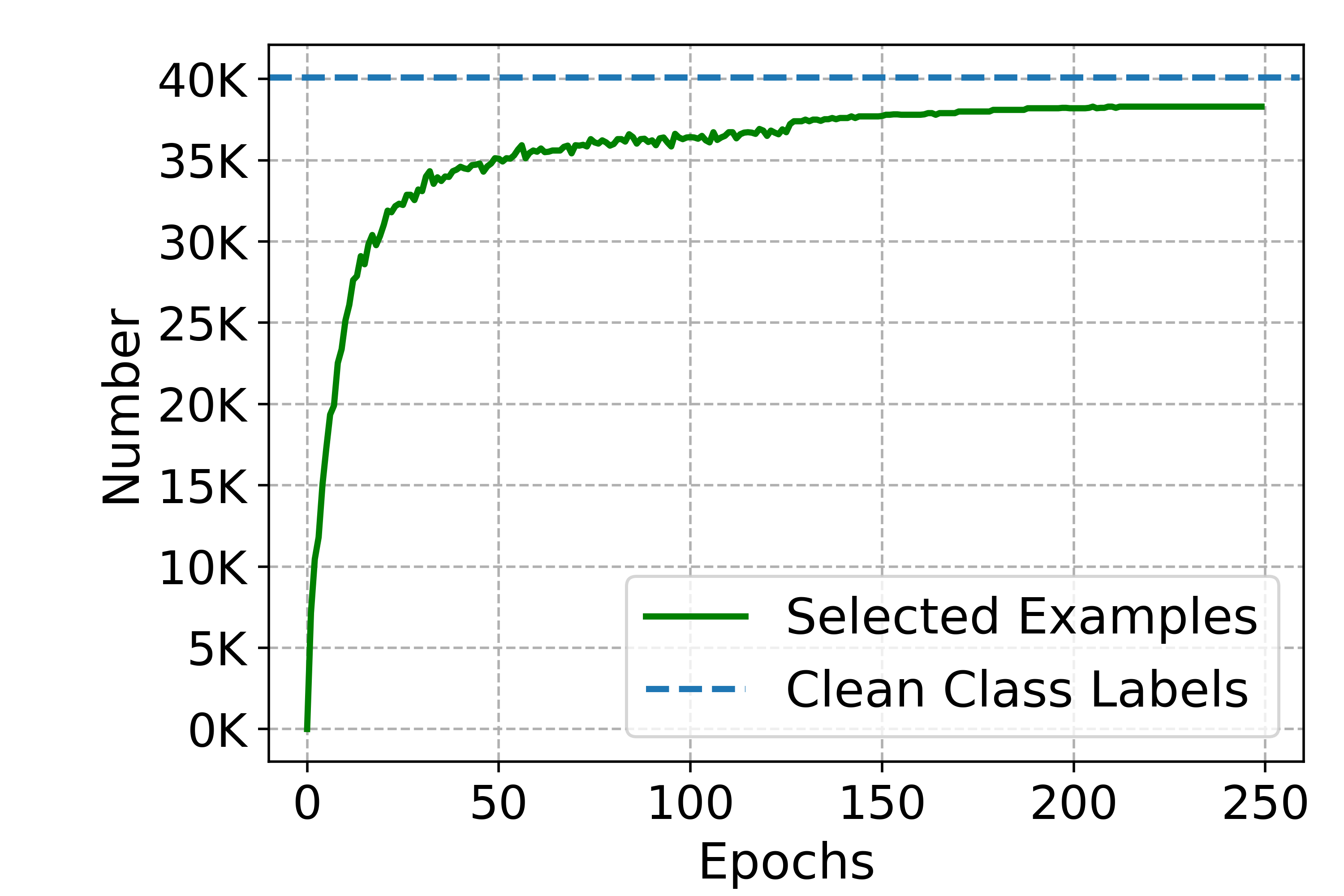}}
	%\hfil
	\subfloat[]{\includegraphics[width=1.65in,trim=30 0 0 0,clip]{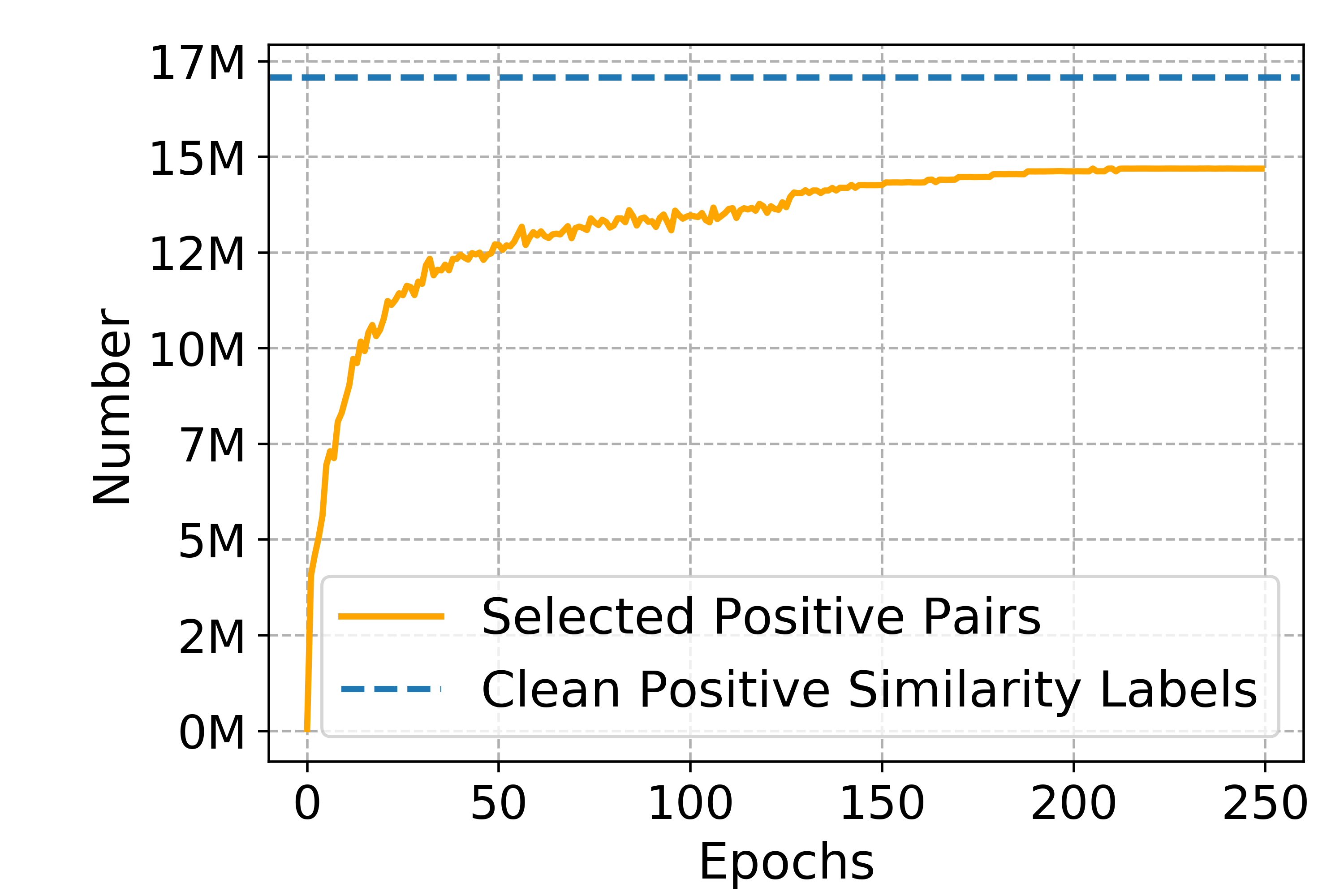}}
	\vspace{6pt}
	\subfloat[]{\includegraphics[width=1.65in,trim=30 0 0 0,clip]{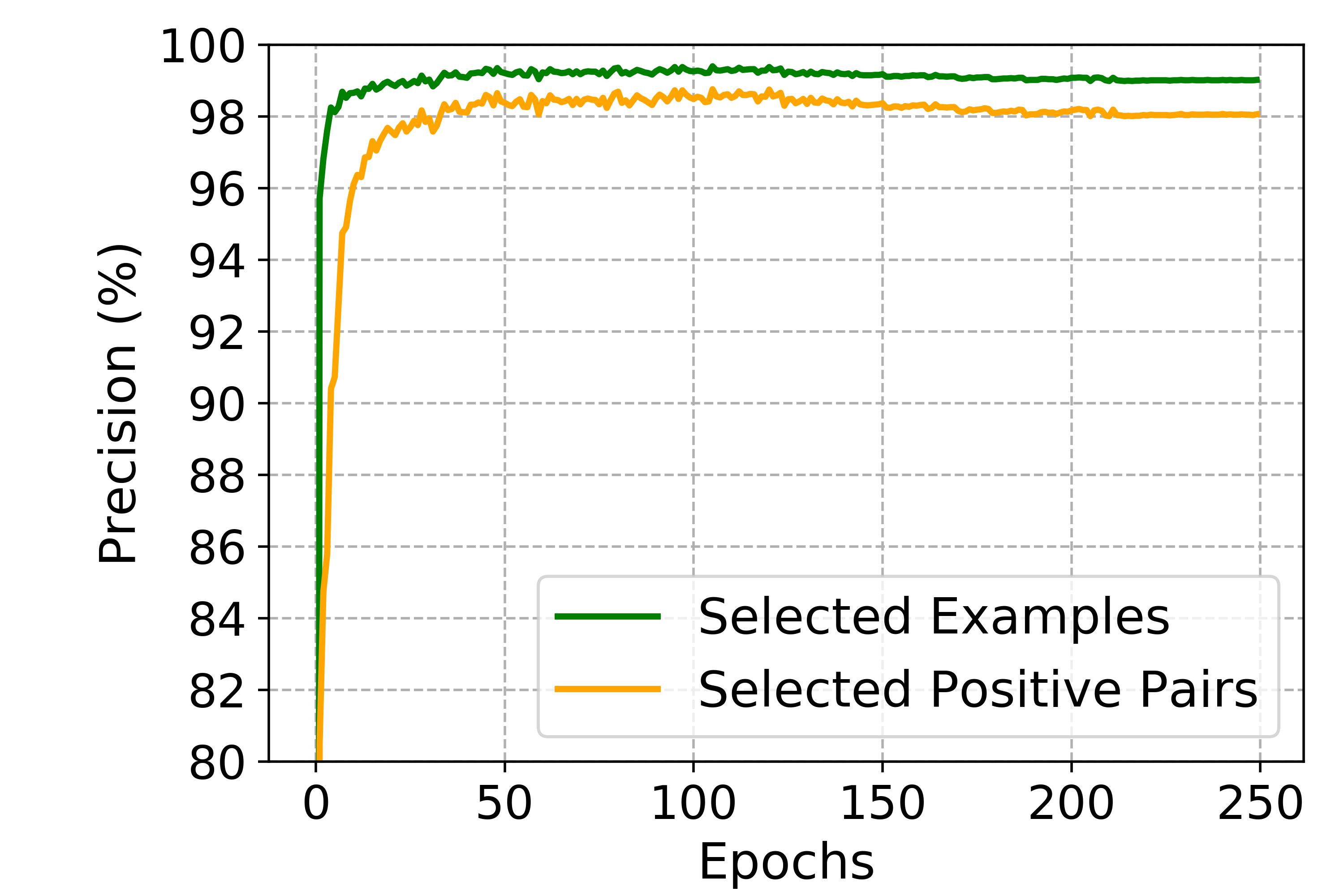}}
	%\hfil
	\subfloat[]{\includegraphics[width=1.65in,trim=30 0 0 0,clip]{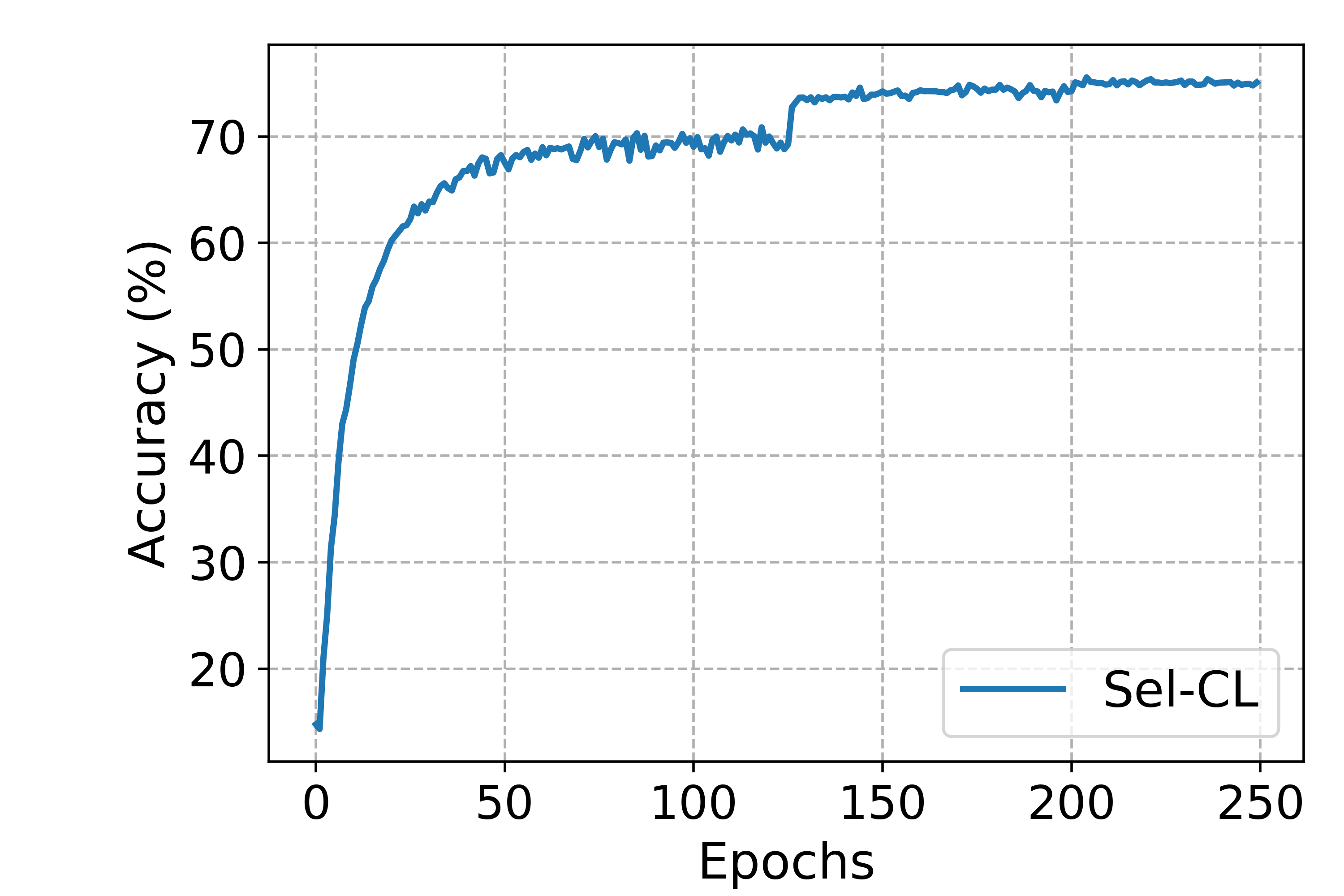}}
	
	\caption{The learning process of Sel-CL on CIFAR-100 with 20\% symmetric noise. (a) the number of selected examples \textit{vs.} epochs; (b) the number of selected positive pairs \textit{vs.} epochs; (c) the label precision of selected examples and pairs (\%) \textit{vs.} epochs; (d) weighted KNN evaluations of Sel-CL (\%) \textit{vs.} epochs. }
	\label{learning_process}
	\vspace{-5pt}
\end{figure}

We start by analyzing the behaviors of different contrastive learning methods in the presence of noisy labels. We show how the selective module for confident pairs impacts the learned representations. We evaluate the quality of representations using a weighted KNN (K=200) evaluation in unsupervised learning~\cite{HuangDGZ19}. 

The results are shown in Tab.~\ref{knn}. As can be seen, when there is no noise, MOIT greatly improves the quality of learned representations. Benifiting from the application of the similarity loss, Sel-CL achieves 0.46\% quality gains. Then, for noisy datasets, Sup-CL is seriously affected by noisy labels. Although MOIT improves its robustness through the Mixup technique and semi-supervised strategy, its performance is still unsatisfactory. For example, in the case with 80\% symmetric noise, the weighted KNN accuracy of MOIT is 55.58\%, which is even less than the result of Uns-CL, \textit{i.e.}, 56.23\%. In contrast, our Sel-CL significantly alleviates the side effect of noisy labels, by selecting confident pairs for supervised contrastive learning. We can see that Sel-CL consistently outperforms baselines across different cases. 

We illustrate the numbers of the selected confident examples and pairs, the label precision of selected confident examples and pairs, and the quality of learned representations during training. The experiments are conducted with 20\% symmetric noise. The results are provided in Fig.~\ref{learning_process}. It can be seen that the selection of confident examples/pairs and representation learning form a positive cycle. Sel-CL can progressively achieve effective selection and improve learned representations.

\subsection{Results on Simulated Noisy Datasets}\label{sec:4.3}

We compare the proposed Sel-CL+ with multiple baselines. We report the averaged test accuracy over the last 10 epochs. All baselines use the PreAct ResNet-18 network. Note that for a fair comparison, we report the results of DivideMix~\cite{LiSH20} without model ensembles, and equip ELR~\cite{LiuNRF20} with Mixup and weight averaging techniques. 

As shown in Tab.~\ref{cifar10}, we first observe that the methods pre-trained by Uns-CL~\cite{chen2020simple} can outperform the original ones, \textit{e.g.}, GCE \textit{vs.} GCE (Uns-CL init.) and ELR \textit{vs.} ELR (Uns-CL init.). When the noise level is high, \textit{e.g.}, 80\% and 90\%, the improvement is clearer, which is consistent with the observations in related
works~\cite{Ghosh2021,Zheltonozhskii2021}. For MOIT+ which exploits Sup-CL by adding regularization, its performance is promising. For symmetric noise, MOIT+ achieves comparable results with other state-of-the-art algorithms, \textit{e.g.}, ELR and DivideMix. For asymmetric noise, MOIT+ can perform better in most cases. As for our Sel-CL+, it achieves competitive performance with symmetric noise. For asymmetric noise, Sel-CL+ consistently achieves the best performance over all baselines. The results verify the effectiveness of our method well.

\subsection{Pair-wise Selection Analysis}\label{sec:4.4}

\begin{table}[t]
	\caption{Weighted KNN evaluation (\%) of Sel-CL with different selected sets on CIFAR-10/100 with 40\% asymmetric noise. }
	\centering
	\footnotesize
	\setlength\tabcolsep{4pt}
	\begin{tabular}{l|c|c}
		\hline
		Sel-CL with different selected sets & CIFAR-10 & CIFAR-100 \\
		\hline  { All examples and all pairs}  & 90.58 & 68.66 \\
		\hline
		{ Confident examples $\mathcal{T}$ and pairs $\mathcal{G}^{\prime}$ }  & 90.64 & 70.25 \\
		{ Confident examples $\mathcal{T}$ and pairs $\mathcal{G}^{\prime} \cup \mathcal{G}^{\prime\prime}$ }  & 92.97 & 72.71 \\
		\hline
		{ Clean examples and associated pairs }  & 94.21 & 71.45 \\
		{ Clean examples and all pairs }    & 94.76 & 73.43 \\		
		{ Clean examples and clean pairs }  & 95.52 & 76.56\\
		\hline
	\end{tabular}
	\label{select}
	\vspace{-5pt}
\end{table}

As the results on simulated noisy datasets show that Sel-CL+ work better in the case of asymmetric noise, where label flips happens between semantically similar classes. It is because Sel-CL not only exploits confident pairs associated with $\mathcal{T}$, but also extracts more confident positive pairs based on the representation similarity. To detail this phenomenon, we conduct some analysis about the pair-wise selection in our approach under the asymmetric noise cases. 

There are several observations provided in Tab.~\ref{select}. The results show that, it is useful to employ the confident examples and associated positive pairs $\mathcal{G}^{\prime}$. However, the imporvement is limited compared with selecting clean examples and clean pairs. Our method further employ $\mathcal{G}^{\prime\prime}$, which is based on the similarity representations. By this procedure, we can make better use of the confident pairs whose class labels are incorrect but similarity labels could be correct, which is much common in the asymmetric cases and real-world scenarios. Besides, an interesting phenomenon is that it is not bad for selecting clean examples and all pairs, since the noise rate of pairs is smaller. This phenomenon may explain why MOIT+ also has advantages with asymmetric noise, which uses all noisy pairs.

%\vspace{-1pt}
\subsection{Results on the Real-World Noisy Dataset}\label{sec:4.5}
We validate our method on the real-world
dataset WebVision-50~\cite{abs_1708_02862}, which contains noisily-labeled images collected from Flickr and Google. As shown in Tab.~\ref{webvision}, Sel-CL+ achieves the best results on both top-1 and top-5 accuracy on the WebVision validation set and ImageNet ILSVRC12 validation set than the other state-of-the-art methods, including four recent methods that also utilize contrastive learning. The results means that our method is better to handle realistic scenes. 
%\vspace{-1pt}

\begin{table}[h]		
	\caption{Accuracy (\%) on the WebVision and ILSVRC2012 validation sets. The model is trained on WebVision-50. The best results are in \textbf{bold}.}\vspace{-5pt}% {*} denotes results acquired by us based on published code.Results of other baselines are from~\cite{LiSH20,Zhi2021ICCV}.
	\centering
	\small
	\begin{tabular}{l|c|c|c|c}
		\hline \multirow{2}{*}{ Methods } & \multicolumn{2}{|c|}{ WebVision } & \multicolumn{2}{c}{ ILSVRC12 } \\
		\cline { 2 - 5 } & top-1 & top-5 & top-1 & top-5 \\
		\hline Forward~\cite{PatriniRMNQ17} &  61.12  &  82.68  &  57.36  &  82.36  \\
		Decoupling~\cite{malach2017decoupling} &  62.54  &  84.74  &  58.26  &  82.26  \\
		D2L~\cite{MaWHZEXWB18} &  62.68  &  84.00  &  57.80  &  81.36  \\
		MentorNet~\cite{Jiang2018icml} &  63.00  &  81.40  &  57.80  &  79.92  \\
		Co-teaching~\cite{Han2018NIPS} &  63.58  &  85.20  &  61.48  &  84.70  \\
		Iterative-CV~\cite{ChenLCZ19} &  65.24  &  85.34  &  61.60  &  84.98  \\
		DivideMix~\cite{LiSH20} &  77.32  &  91.64  & 75.20 &  90.84  \\
		ELR~\cite{LiuNRF20} &  76.26  &  91.26  &  68.71  &  87.84  \\
		ELR+~\cite{LiuNRF20} &  77.78  &  91.68  &  70.29  &  89.76  \\
		\hline 
		\underline{ELR (Uns-CL init.)} & 79.93 & 92.00 & 71.23 & 88.23\\
		ProtoMix~\cite{li2020learning} & 76.3  & 91.5  & 73.3  & 91.2 \\
		MoPro~\cite{Li2021ICLR}  & 77.59 & -- & 76.31 & -- \\
		NGC~\cite{Zhi2021ICCV}   & 79.16 & 91.84 & 74.44 & 91.04 \\
		Sel-CL+ & \textbf{79.96} & \textbf{92.64} & \textbf{76.84} & \textbf{93.04} 
		\\
		\hline
	\end{tabular}
    \label{webvision}
\end{table}
	
\subsection{Ablation Study and Discussions} \label{sec:4.6}
\vspace{-2pt}
\myPara{Influence of each component.}
To study the impact of each component in our method, we use CIFAR-100 for evaluations. We report test accuracy/weighted KNN evaluations~(\%) for Sel-CL, and test accuracy~(\%) for Sel-CL+ in Tab.~\ref{ablation}. It shows the contribution of each component into our method. 
\begin{table}[t]		
	\caption{Ablation study for Sel-CL and Sel-CL+ on CIFAR-100. The best results are in \textbf{bold}.}\vspace{-5pt}
	\centering
	\small
	\begin{tabular}{l|c|c}
		\hline \multirow{1}{*}{Methods}  &\multicolumn{1}{c}{ Sym. 20\%} & \multicolumn{1}{|c}{Asym. 40\%} \\
		\hline
		Sel-CL w/o Mixup Data Aug. & 70.3/70.6 & 64.2/66.2  \\
		Sel-CL w/o MOCO Trick & 73.3/74.1 &  69.2/71.5 \\
		Sel-CL w/o Selection  & 67.2/68.9 & 49.9/68.7  \\		
		%Sel-CL w/o Selecting $\mathcal{G}^{\prime\prime}$  & xx.x/xx.x & xx.x/70.3  \\
		Sel-CL w/o Classfier Learning & ~~---~~/69.9\quad & ~~---~~/70.2\quad \\		
		Sel-CL w/o $\mathcal{L}^{SIM}$  & 74.5/74.9 & 71.8/72.5  \\
		Sel-CL & \textbf{74.9/75.4} & \textbf{72.0/72.7}  \\
		\hline
		Sel-CL+ w/ Strong Data Aug. & 74.5 & 72.7  \\
		Sel-CL+ w/o Retraining Cls. & 76.4 & 73.4  \\
		Sel-CL+ & \textbf{76.5} & \textbf{74.2} \\				
		\hline
	\end{tabular}
    \label{ablation}
   \vspace{-6pt}
\end{table}

\myPara{Discussions on warm-up methods.} We use different warm-up methods, \textit{i.e.}, Uns-CL~\cite{chen2020simple} and Sup-CL~\cite{Khosla2020} for Sel-CL+. The experiments are conducted on CIFAR-10 and CIFAR-100 with symmetric and asymmetric noise. The results in Tab.~\ref{warmup} demonstrate that Sel-CL+ is robust to the choice of warm-up methods.

\myPara{Discussions on fine-tuning methods.}
We further test two complex but advanced methods for fine-tuning stage, \textit{i.e.}, DivideMix~\cite{LiSH20} (be better with symmetric noise) and ELR+~\cite{LiuNRF20} (be better with asymmetric noise). As shown in Tab.~\ref{finetuning}, using more advanced fine-tuning methods can further improve the performance. Both the representations obtained by Uns-CL~\cite{chen2020simple} and Sel-CL can promote the robustness of two methods, and our Sel-CL brings better results.

\begin{table}[h]
	\caption{Comparison with different warm-up methods in the test accuracy (\%) of Sel-CL+.} \vspace{-5pt}
	\centering
	\small
	\setlength\tabcolsep{5pt}
	\begin{tabular}{l|c|c|c|c|c|c}
		\hline \multirow{1}{*}{Dataset} & \multicolumn{3}{c}{CIFAR-10} & \multicolumn{3}{|c}{CIFAR-100} \\
		\hline \multirow{1}{*}{Noise type} & \multicolumn{2}{c}{Sym.} & \multicolumn{1}{|c}{Asym.} & \multicolumn{2}{|c}{Sym.} & \multicolumn{1}{|c}{Asym.}\\
		\hline
		Noise rate & 
		20\% & 90\% & 40\% & 20\% & 90\% & 40\% \\
		\hline
		Uns-CL~\cite{chen2020simple} & 95.5 & 81.9 & 93.4 & 76.5 & 48.8 & 74.2   \\
		Sup-CL~\cite{Khosla2020} & 95.5 & 81.6 & 93.4 & 76.8 & 51.4  & 74.5 \\
		\hline
	\end{tabular}
	\label{warmup}
\end{table}

\begin{table}[h]
	\caption{Comparison with using different fine-tuning methods in the test accuracy (\%). The best results are in \textbf{bold}.}\vspace{-5pt} %{*} denotes results acquired by us based on published code. Results of other baselines are from~\cite{Zheltonozhskii2021,LiuNRF20}. 
	\centering
	\small
	\setlength\tabcolsep{4.5pt}
	\begin{tabular}{l|c|c|c|c}
		\hline \multirow{1}{*}{Dataset} & \multicolumn{2}{c}{CIFAR-10} & \multicolumn{2}{|c}{CIFAR-100} \\
\hline \multirow{1}{*}{Noise type} & \multicolumn{1}{c}{Sym.} & \multicolumn{1}{|c}{Asym.} & \multicolumn{1}{|c}{Sym.} & \multicolumn{1}{|c}{Asym.}\\
\hline
		Noise rate & 
		20\%  & 40\% & 20\% & 40\% \\
		\hline			
		DivideMix~\cite{LiSH20} & 95.7 & 92.1 & 76.9 & \underline{53.8}\\		
		ELR+~\cite{LiuNRF20} & 94.6 & \underline{93.0} & 77.5 & \underline{72.2} \\
		\hline
		%ELR (Uns-CL init.){*}	& 94.4 & 93.3 & 76.2 & 70.4 \\						
		DivideMix (Uns-CL init.)~\cite{Zheltonozhskii2021} & 96.2  & 90.8 & 78.3 & \underline{52.9}\\
		\underline{ELR+ (Uns-CL init.)}~\cite{Zheltonozhskii2021} & 94.8 & 94.3 & 77.7 & 72.3 \\
		
		\hline
        %Sel-CL+	& 95.1 & 93.4 & 76.3 & 74.2 \\			
		DivideMix (Sel-CL init.) & \textbf{96.3} & 91.6 & \textbf{78.7} & 55.2 \\
		ELR+ (Sel-CL init.) & 95.2 & \textbf{94.6} & 77.7 &  \textbf{72.9} \\
		\hline
	\end{tabular}
	\label{finetuning}
\end{table}

\begin{table}[t]
	\caption{Comparesion with one-stage methods in the test accuracy (\%). $\dagger$ denotes fine-tuning using AugMix data augmentation. The best results are in \textbf{bold}.} 
	\vspace{-5pt}%{*} denotes results acquired by us based on published code. $\dagger$ denotes fine-tuning using AugMix data augmentation. %Results of other baselines are from~\cite{Zhi2021ICCV}.
	\centering
	\small
	\setlength\tabcolsep{5pt}
	\begin{tabular}{l|c|c|c|c|c|c}
		\hline \multirow{1}{*}{Dataset} & \multicolumn{3}{c}{CIFAR-10} & \multicolumn{3}{|c}{CIFAR-100} \\
		\hline \multirow{1}{*}{Noise type} & \multicolumn{2}{c}{Sym.} & \multicolumn{1}{|c}{Asym.} & \multicolumn{2}{|c}{Sym.} & \multicolumn{1}{|c}{Asym.}\\
		\hline
		Noise rate & 
		20\% & 90\% & 40\% & 20\% & 90\% &  40\% \\
		\hline			
		ProtoMix~\cite{li2020learning} & 95.8 & 75.0 & 91.9 & 79.1 & 29.3 & \underline{48.8} \\		
		NGC~\cite{Zhi2021ICCV} & \textbf{95.9} & \textbf{80.5} & 90.6 & \textbf{79.3} & 29.8 & --\\
		\hline
		%Sel-CL+ & 95.1 & 83.1 & 93.4 & 76.3 & 51.3 & 74.2 \\
		%Sel-CL & 94.5 & 67.4 & 91.9 & 76.7 & 32.6 & \textbf{74.3} \\			
		Sel-CL+ & 95.4 & 67.5 & \textbf{92.8} & 76.4 & \textbf{35.5} &  \textbf{74.2} \\
		Sel-CL+$^\dagger$ & 95.2 & 67.4 & 92.5 & 76.0 & 35.4 & \textbf{74.2} \\
		\hline
	\end{tabular}
	\label{one-stage}
	%\vspace{-6pt}
\end{table}

\myPara{Comparison with one-stage methods for handling noisy labels.}	
We compare our method with two recent one-stage methods, which also employ contrastive learning to handle noisy labels. For a fair comparison, we use AugMix~\cite{HendrycksMCZGL20} as data augmentations of the pre-training stage. As shown in Tab.~\ref{one-stage}, our approach has advantages in the asymmetric noise cases. In addition, we also try to adopt AugMix augmentation in the fine-tuning stage and find that the weak augmentation is more suitable, which is consistent with the ablation study. The results may reflect the difference between representation learning and classification learning.

\begin{table}[t]
	\caption{Comparison with different example selection strategies in the test accuracy/label precsion (\%) of Sel-CL+.} \vspace{-5pt}
	\centering
	\small
	\setlength\tabcolsep{4pt}
	\begin{tabular}{l|c|c|c|c}
		\hline \multirow{1}{*}{Noise type} & \multicolumn{2}{c}{Sym.} & \multicolumn{2}{|c}{Asym.} \\
		\hline
		Noise rate & 
		20\% & 90\% & 20\% & 40\% \\
		\hline
		w pseudo-labels & 76.5/99.1 & 48.8/62.0 & 77.5/97.5 & 74.2/92.2 \\
		w/o pseudo-labels & 76.5/99.1 & 46.2/55.4 & 76.8/96.6 & 69.4/83.9 \\
		\hline
	\end{tabular}
	\label{pseudo-labels}
	\vspace{-6pt}
\end{table}

\begin{figure}[ht]
	\centering
	\includegraphics[width=5.3cm,height=3.35cm]{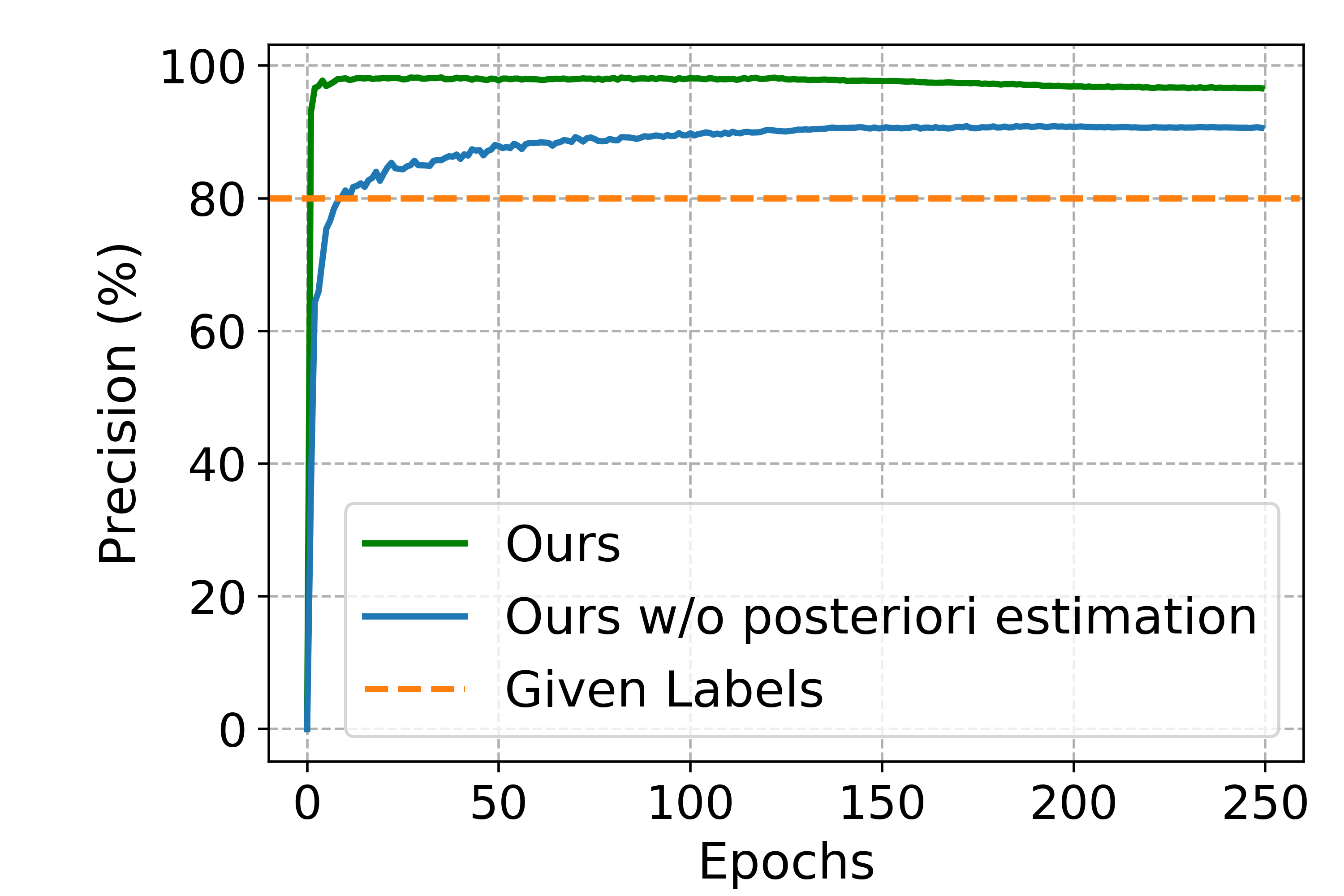}
	\vspace{-3pt}
	\caption{Comparison of label precision. The experiments are conducted on CIFAR-100 with 20\% asymmetric noise.}
	\vspace{-3pt}
	\label{fig:precision}
\end{figure}

\myPara{Role of pseudo-labels $\hat{y}$.} Following~\cite{Ortego2021}, we use the representation similarity with the weighted KNN to create pseudo labels.  In this way, we can make better use of representation similarities. To make it clear, we use CIFAR-100 for evaluations. We report test accuracy/label precsion
evaluations~(\%) for Sel-CL+ with and without pseudo-labels, and here label precsion is the average over all epochs. 
As Tab.~\ref{pseudo-labels} shows, this practice results in better performance due to improved label precision. Besides, we also can just use the pseudo labels to define confident examples rather than to estimate clean class posterior probabilities. By comparing them from empirical observations, our method brings a higher label precision, which is shown in Fig.~\ref{fig:precision}. %intuitively

The ablation studies about the hyperparameters are provided in Appendix A, which show that our method is robust to the choices of the hyperparameters.
\vspace{-3pt}
\section{Limitations}
Our work still has certain limitations, including: 
1) This work exploits contrastive learning, and therefore,
the performance of our approach relies on adequate data
augmentation and large amounts of negative samples. A
larger batch size or memory bank is needed, which places
higher demands on the storage of computing devices.
2) The use of the KNN algorithm brings greater computational consumption. In this work, we have used some
faster KNN algorithms (see source codes) to alleviate the
above issue, which facilitates the application of our method
to large-scale datasets.

\vspace{-5pt}
\section{Conclusion}\label{sec:5}
This paper proposes selective-supervised contrastive learning (Sel-CL), a new method to handle noisy labels in training data by learning robust pre-trained representations. We make use of the pair-wise characteristic of contrastive learning to better enhance network robustness. Without the noise rate prior, the confident pairs are selected out of noisy pairs for supervised contrastive learning. We demonstrate the state-of-the-art performance of our method with extensive experiments on multiple noisy datasets. For future work, we are interested in extending our method to other tasks such as object detection and text matching.

\small{\myPara{Acknowledgements}}
%\section*{Acknowledgements}
This work was partially supported by grants from the Beijing Natural Science Foundation (19L2040), National Key Research and Development Plan (2020AAA0140001), and National Natural Science Foundation of China (61772513). Shiming Ge was supported by the Youth Innovation Promotion Association, CAS. Xiaobo Xia was supported by ARC Projects DE-190101473. Tongliang Liu was partially supported by ARC Projects DP180103424, DE-190101473, and DP-220102121.

%%%%%%%%% REFERENCES
{\small
\bibliographystyle{ieee_fullname}
\bibliography{bibSelCon}
}

\end{document}

%\title{Supplementary Materials}
%\maketitle

%%%%%%%%% BODY TEXT

\section{Hyperparameter Sensitivity Analysis}
\myPara{Analysis of $\lambda_s$.}
$\lambda_s$ is the balancing weight of a added similarity loss $L^{SIM}$. In the paper, we set it as 0.01 for all experiments. As shown in Tab.~\ref{lambda_s},  our approach is robust to selection of $\lambda_s$.
\begin{table}[h]
	\caption{Test accuracy (\%) of Sel-CL+ with different $\lambda_s$ on CIFAR-10 with 20\% symmetric label noise. }
	\centering
	\small
	
	\begin{tabular}{c|c|c|c|c|c|c}
		\hline
		Parameter  & 0.1 & 0.05 & 0.01 & 0.005 & 0.001 & 0.0001 \\
		\hline  
		$\lambda_s$  & 95.3 & 95.6 & 95.5 & 95.5 & 95.6 & 95.2 \\	
		\hline
	\end{tabular}
	\label{lambda_s}
\end{table}

\myPara{Analysis of $\alpha$ and $\beta$.}
Noise detection fractiles $\alpha$ and $\beta$ are used to determine the dynamic thresholds for selecting confident examples and pairs. We use CIFAR-10 datasets for analysis. In the paper, we set $\alpha=50\%, \beta=25\%$ for CIFAR-10 with simulated label noise. As shown in Tab.~\ref{sym} and Tab.~\ref{asym}, our approach is robust to choices of noise detection fractiles.

\begin{table}[h]
	\caption{Test accuracy (\%) of Sel-CL+ with different $\alpha$ and $\beta$ on CIFAR-10 with 20\% symmetric label noise. }
	\centering
	\small
	\begin{tabular}{c|c|c|c|c|c|c}
		\hline
		Parameter  & 0\% & 15\% & 25\% & 35\% & 50\% & 75\% \\
		\hline  
		$\alpha$ & 94.6 & 94.7 & 95.3 & 95.2 & 95.5 & 95.6  \\
		\hline
		$\beta$  & 94.9 & 95.4 & 95.5 & 95.5 & 95.0 & 95.1\\	
		\hline
	\end{tabular}
	\label{sym}
\end{table}

\begin{table}[h]
	\caption{Test accuracy (\%) of Sel-CL+ with different $\alpha$ and $\beta$ on CIFAR-10 with 40\% asymmetric label noise. }
	\centering
	\small
	\begin{tabular}{c|c|c|c|c|c|c}
		\hline
		Parameter  & 0\% & 15\% & 25\% & 35\% & 50\% & 75\% \\
		\hline  
		$\alpha$  & 92.1 & 92.5 & 92.6 & 93.0 & 93.4 & 89.9\\
		\hline
		$\beta$  & 91.4 & 93.3 & 93.4 & 92.6 & 92.8 & 92.5\\	
		\hline
	\end{tabular}
	\label{asym}
\end{table}

\section{Implementation Details For Baselines}
For a fair comparison, except for the results borrowed from related work, we obtain baselines based on published codes with the recommended or well-tuned hyperparameters. The details are as following.

\myPara{Uns-CL} For CIFAR-10/100 datasets, we train the network for 1000 epochs with SimAug augmentation and cosine learning rate, where the scalar
temperature is 0.07, the batch size is 1024 and the initial learning rate is 0.05. For WebVison-50 dataset, we use the pre-trained ResNet-50 model provided by the source code of C2D method~\footnote{https://github.com/ContrastToDivide/C2D}, which is trained for 1000 epochs using SimCLR. 
%~\footnote{https://github.com/HobbitLong/SupContrast}
%~\footnote{https://github.com/ContrastToDivide/C2D}

\myPara{GCE} For CIFAR-10 datasets, we train the networks for 120 epochs with $q=0.5$ for 20\%/50\% sym. noise, $q=0.7$ for 80\%/90\% sym. noise, and  $q=0.5$ for asym. noise. For CIFAR-100 datasets, we train the networks for 150 epochs with $q=0.3$.

\myPara{GCE (Uns-CL init.)} For CIFAR-10 datasets, we train the networks for 120 epochs with $q=0.7$ for 20\% sym. noise, $q=1.0$ for 50\%/80\%/90\% sym. noise, and  $q=0.5$ for asym. noise. For CIFAR-100 datasets, we train the networks for 150 epochs with $q=0.7$ for sym. noise and  $q=0.5$ for asym. noise.

\myPara{ELR} We train the networks for 250 epochs with $\lambda=3$ and $\beta=0.7$ for CIFAR-10 with sym. noise, $\lambda=1$ and $\beta=0.9$ for CIFAR-10 with asym. noise, and $\lambda=7$ and $\beta=0.9$ for CIFAR-100.

\myPara{ELR (Uns-CL init.)} We train the networks for 250 epochs with $\lambda=5$ and $\beta=0.7$ for CIFAR-10 with 20\%/50\% sym. noise, $\lambda=7$ and $\beta=0.7$ for CIFAR-10 with 80\%/90\% sym. noise, $\lambda=3$ and $\beta=0.9$ for CIFAR-10 with asym. noise, $\lambda=7$ and $\beta=0.9$ for CIFAR-100 with 20\%/50\% sym. noise, $\lambda=10$ and $\beta=0.9$ for CIFAR-100 with 80\%/90\% sym. noise and asym. noise.

\myPara{MOIT+} For pre-training stage, we train the networks with recommended hyperparameter setting in the original paper, except for CIFAR-100 dataset without label noise, where we do not apply the semi-supervised strategy. For fine-tuning stage, we refine it with recommended setting.

\myPara{DivideMix (Uns-CL init.)} We use the implementation of C2D method to train the networks for with its recommended
hyperparameter setting.

\myPara{ELR+ (Uns-CL init.)} We train the networks for 250 epochs using the same $\lambda$ and $\beta$ with ELR (Uns-CL init.).

\myPara{ProtoMix} For CIFAR-100 with asymmetric label noise, we train the network for 300 epochs with 0.02 initial learning rate, 128 batch size, 0.4 $\eta_{0}$ and 0.9 $\eta_{1}$.

\myPara{Other baselines} We obtain other reproduced baselines with their recommended or default hyperparameters, which include Cross-Entropy, Mixup, Forward, P-correction, M-correction and DivideMix.

\begin{figure*}[!htbp]
	\begin{minipage}[c]{0.05\columnwidth}\centering\small \rotatebox[origin=c]{90}{-- \footnotesize{Cross-Entropy} --} \end{minipage}%
	\begin{minipage}[c]{0.95\textwidth}
		\includegraphics[width=0.2\textwidth]{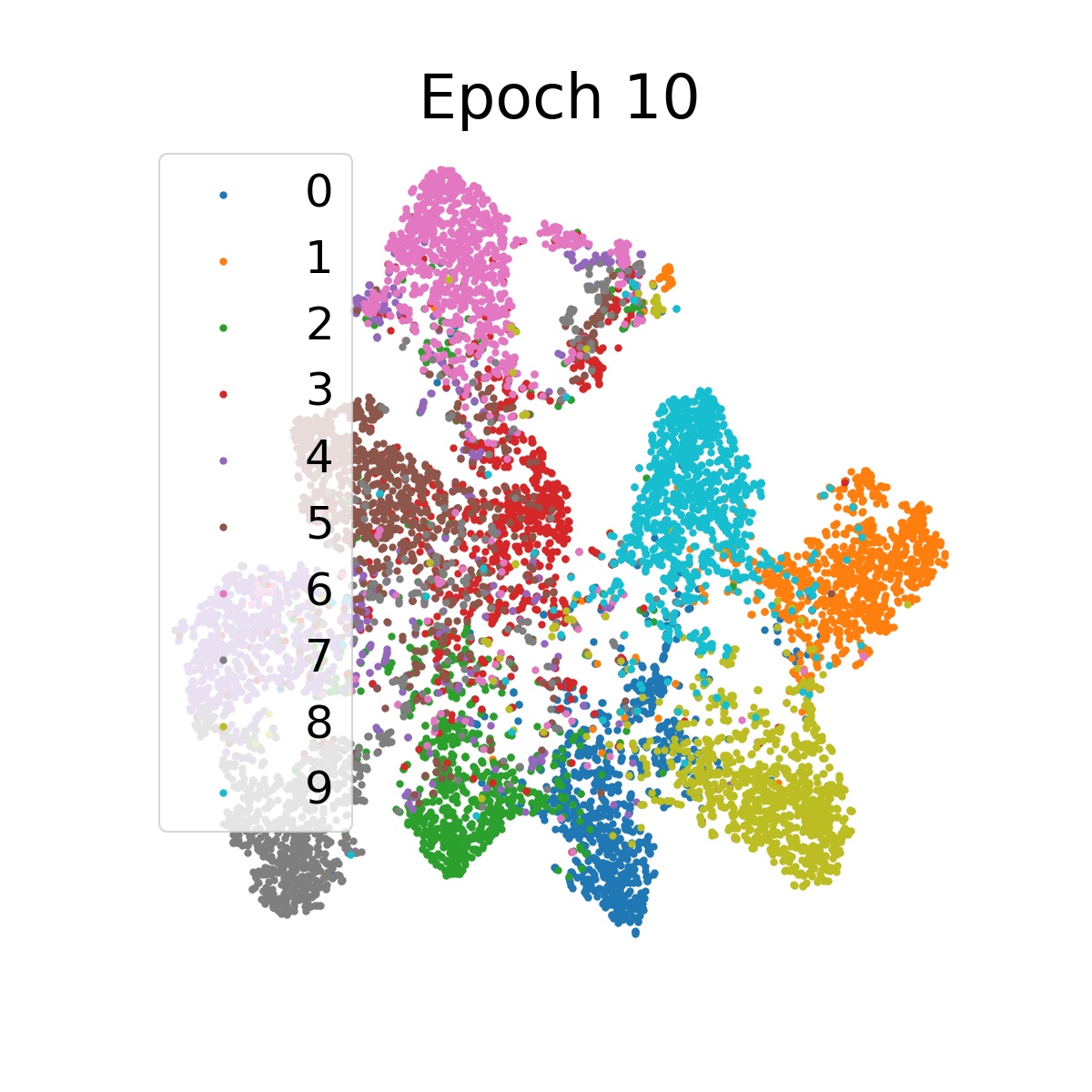}%
		\includegraphics[width=0.2\textwidth]{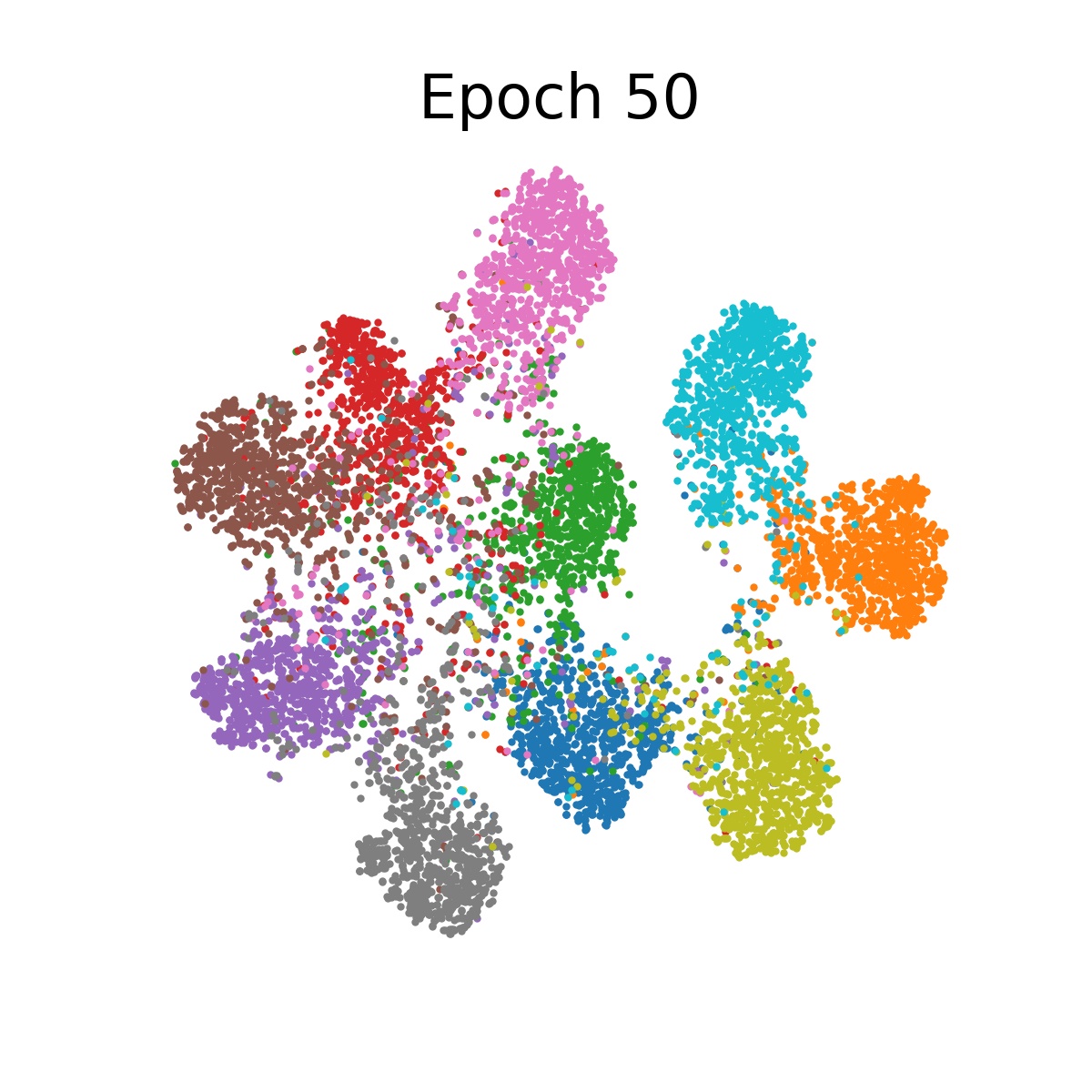}%
		\includegraphics[width=0.2\textwidth]{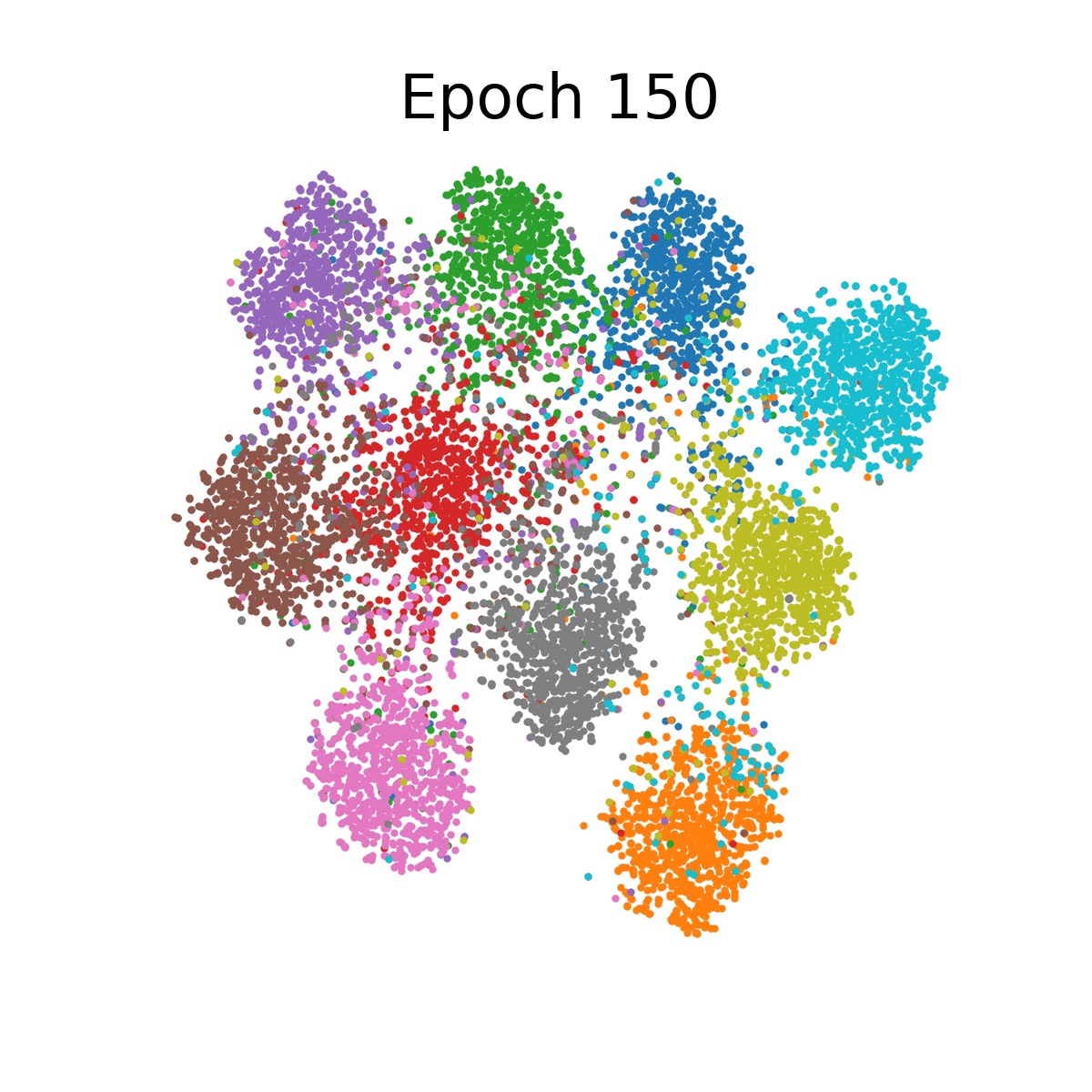}%
		\includegraphics[width=0.2\textwidth]{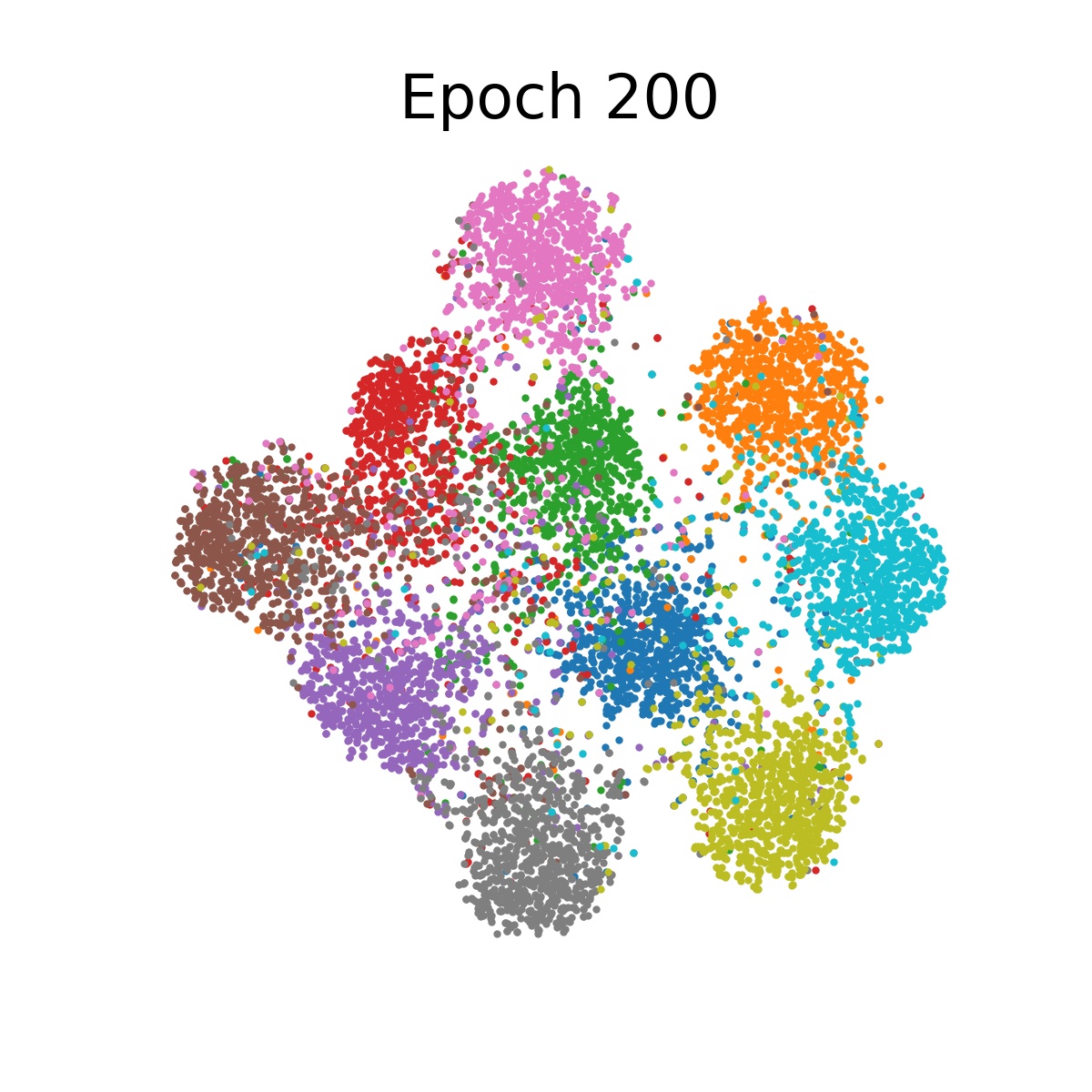}%
		\includegraphics[width=0.2\textwidth]{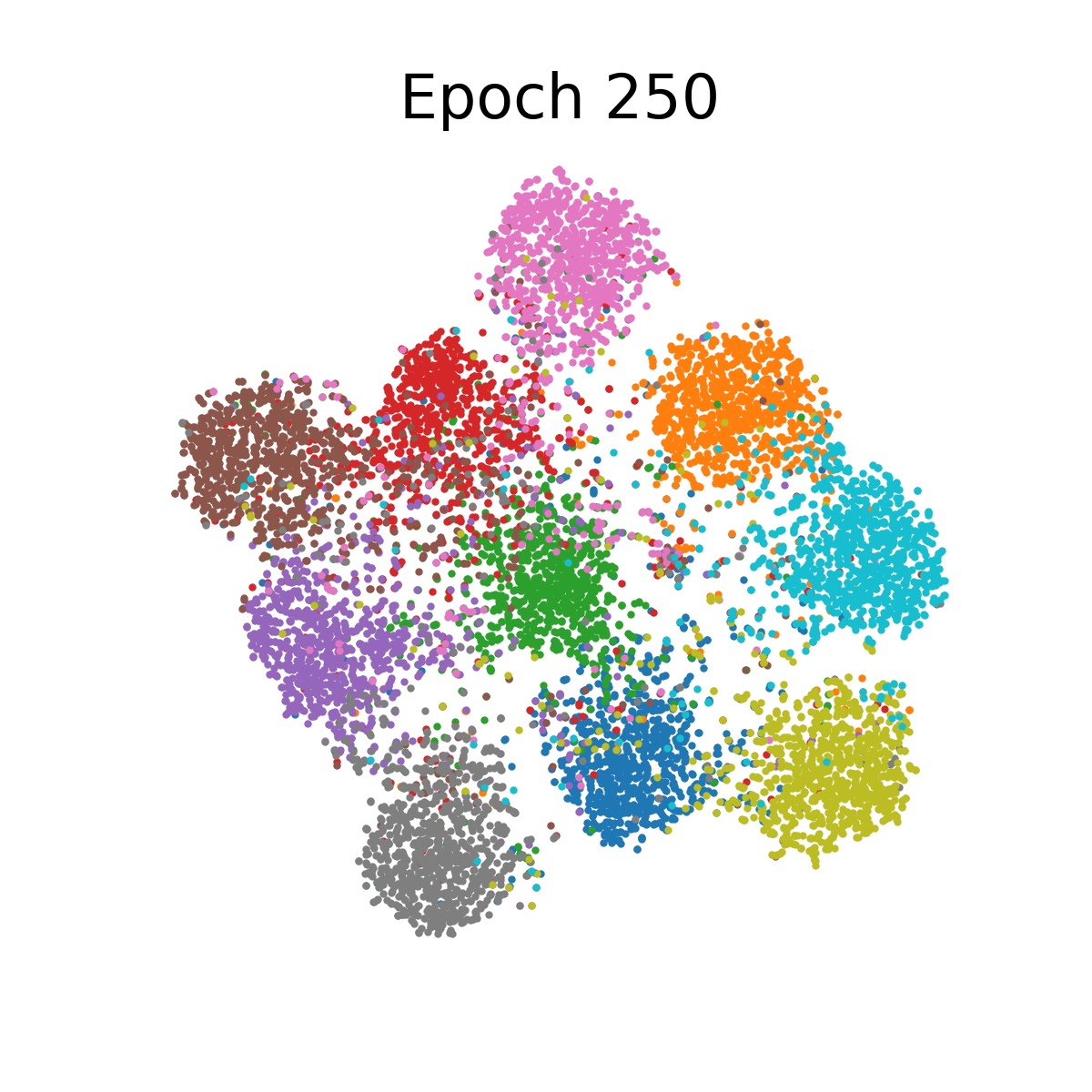}%
	\end{minipage}\vspace{-12pt}
	\begin{minipage}[c]{0.05\columnwidth}\centering\small \rotatebox[origin=c]{90}{-- \footnotesize{Sel-CL} --} \end{minipage}%
	\begin{minipage}[c]{0.95\textwidth}
		\includegraphics[width=0.2\textwidth]{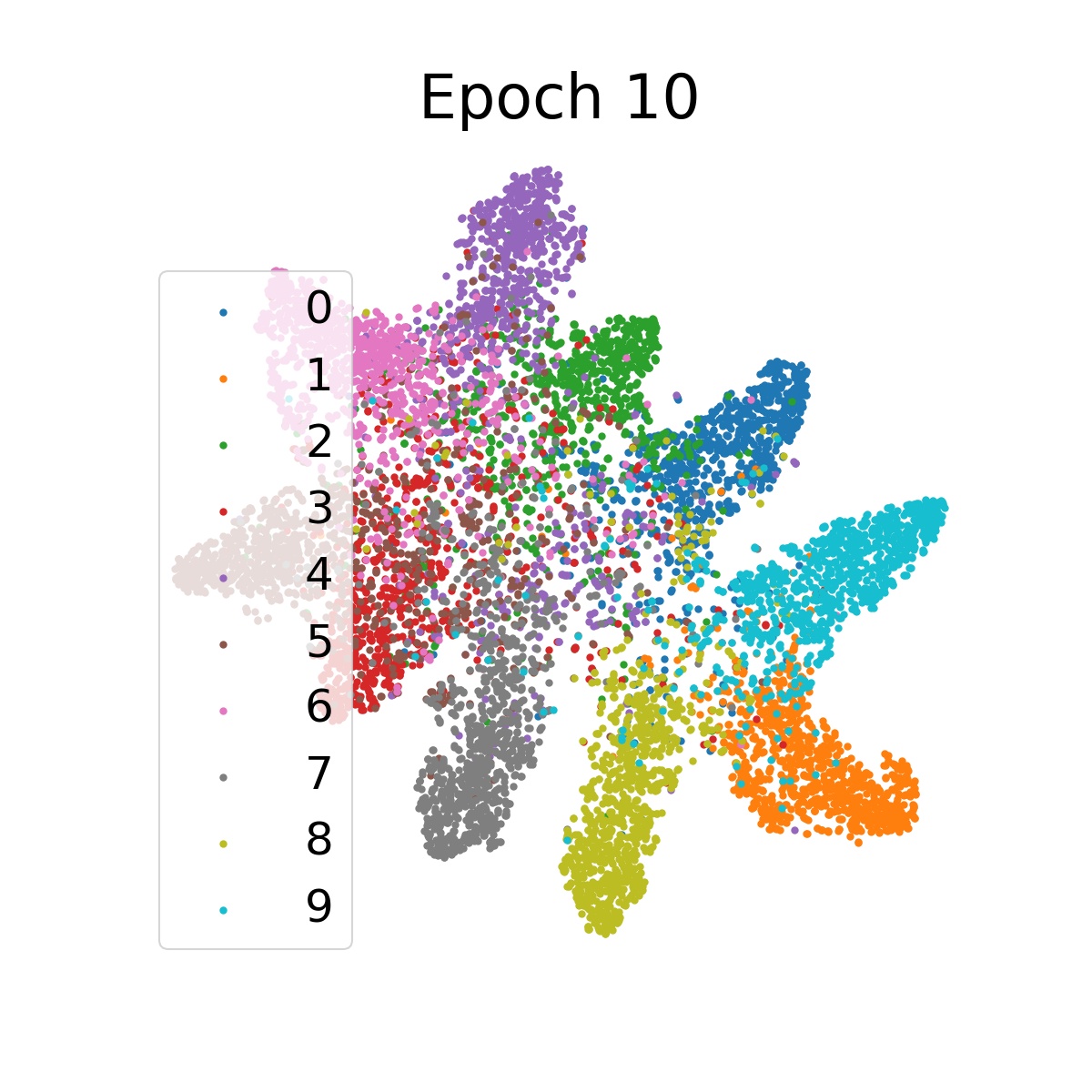}%
		\includegraphics[width=0.2\textwidth]{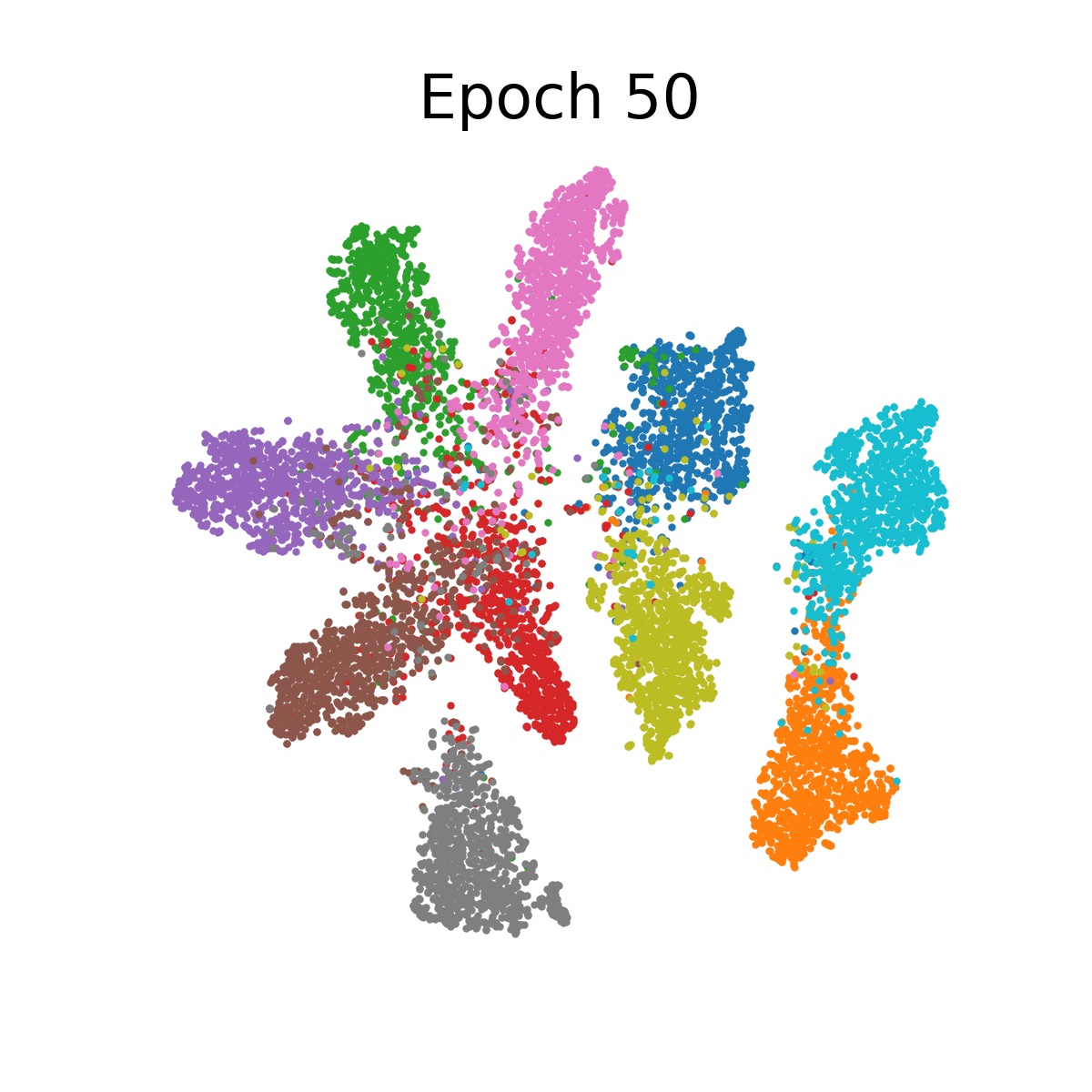}%
		\includegraphics[width=0.2\textwidth]{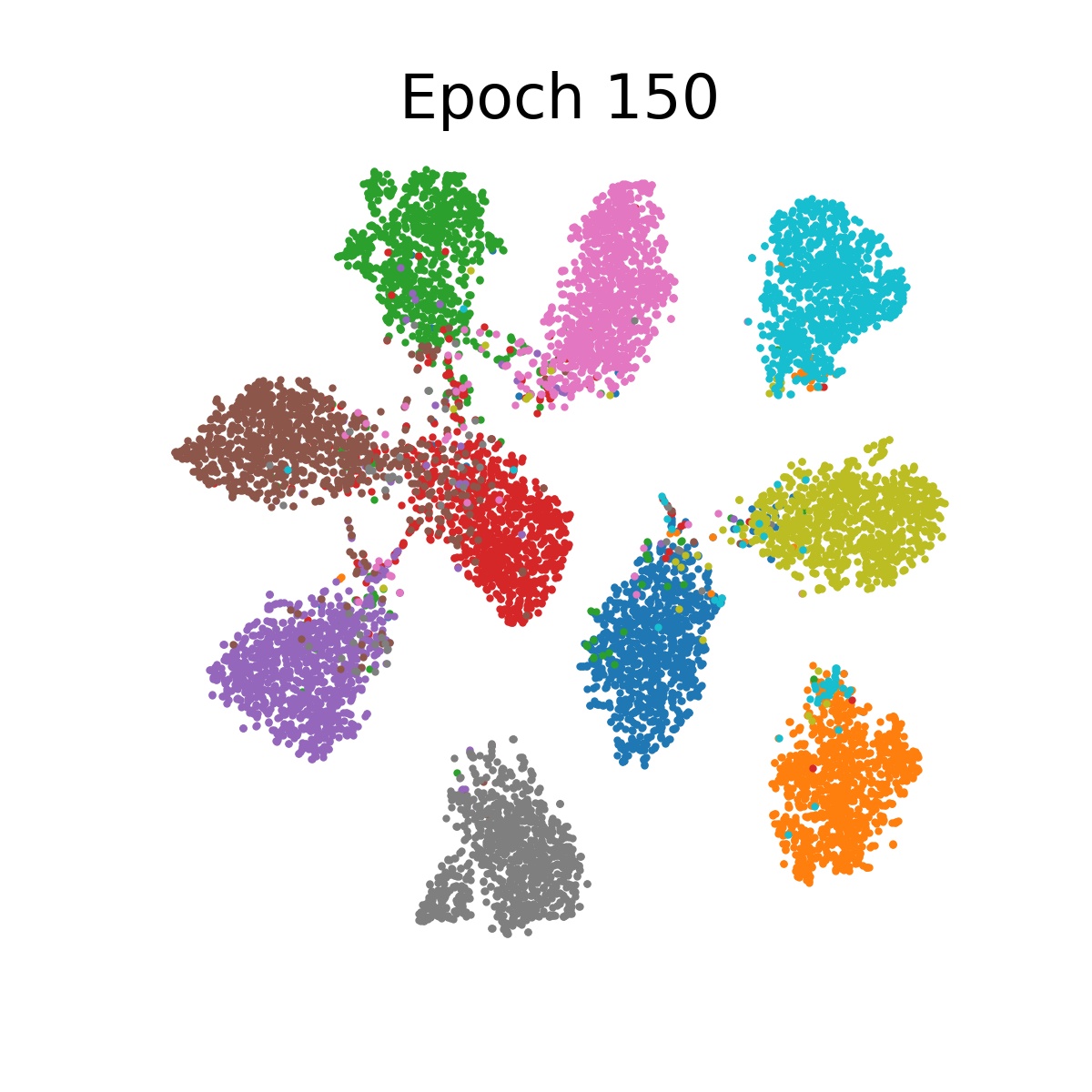}%
		\includegraphics[width=0.2\textwidth]{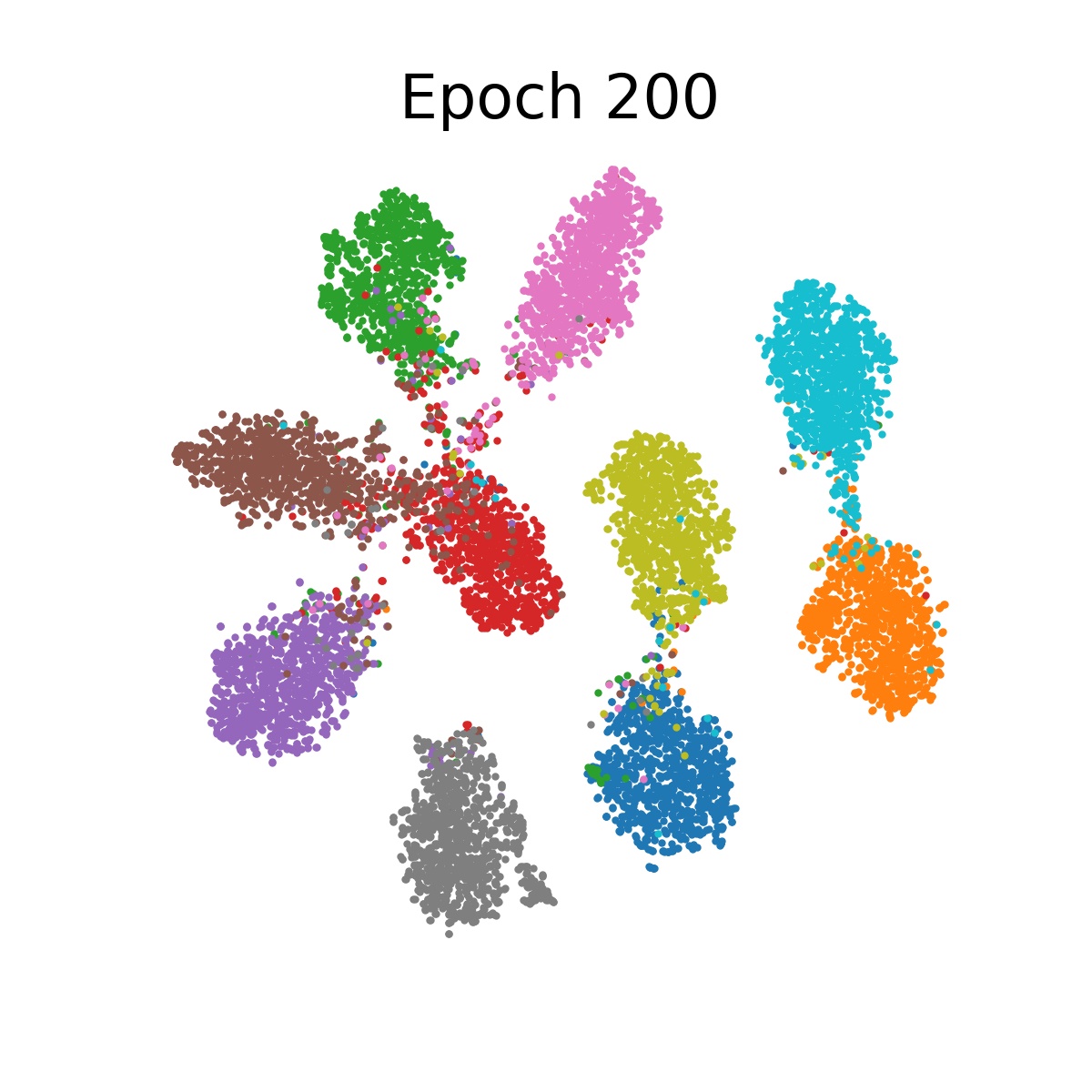}%
		\includegraphics[width=0.2\textwidth]{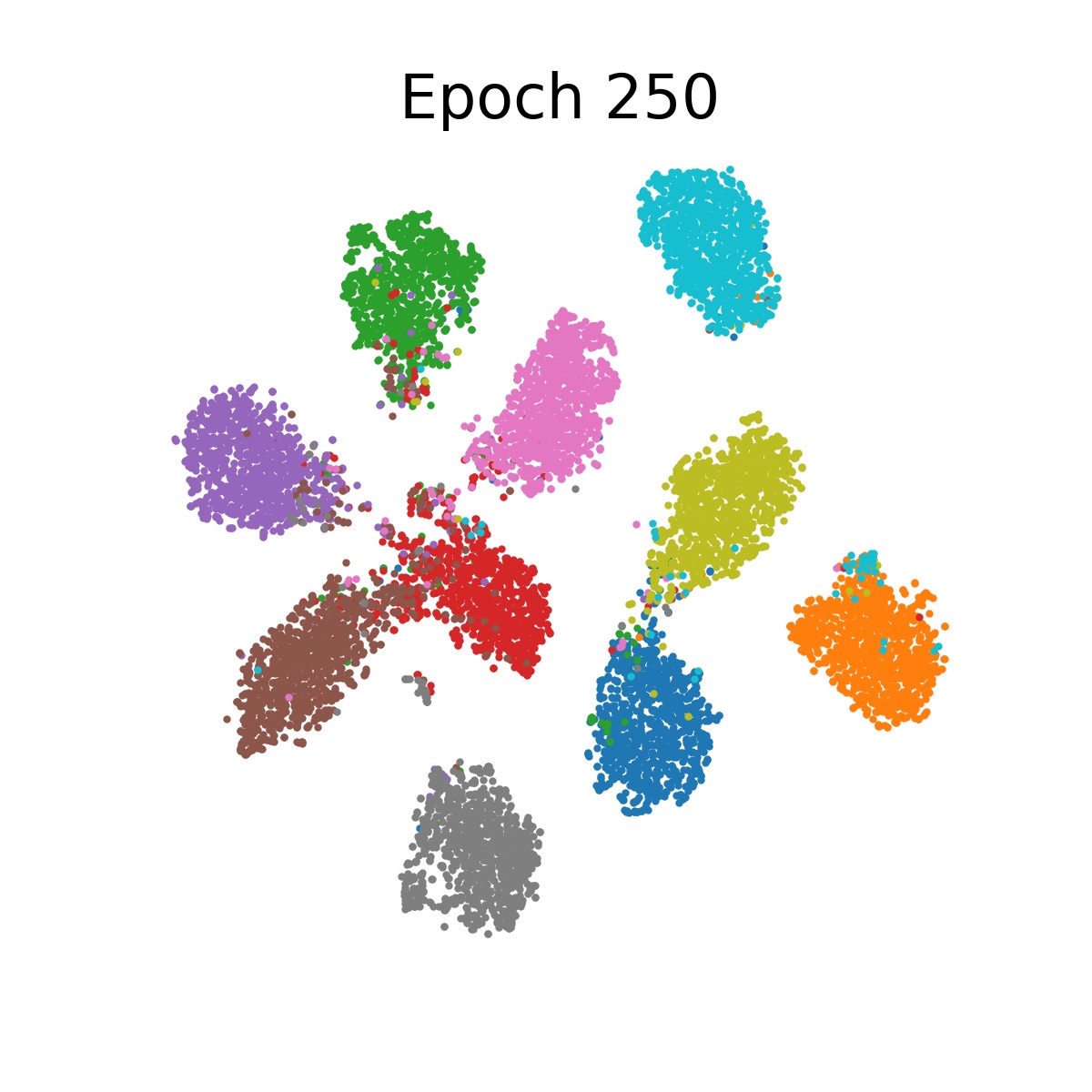}%
	\end{minipage}\vspace{-12pt}
	\begin{minipage}[c]{0.05\columnwidth}\centering\small \rotatebox[origin=c]{90}{-- \footnotesize{Cross-Entropy} --} \end{minipage}%
	\begin{minipage}[c]{0.95\textwidth}
		\includegraphics[width=0.2\textwidth]{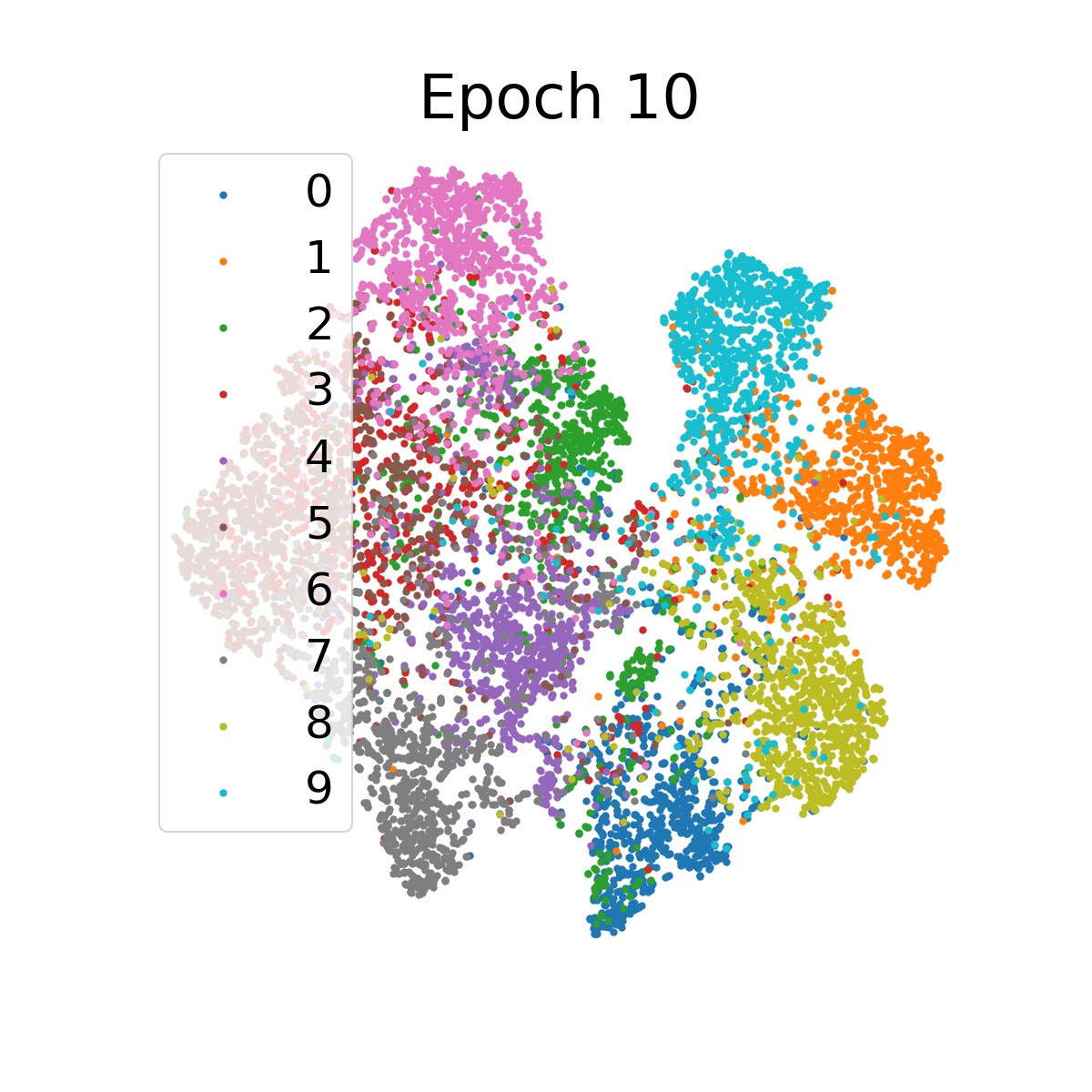}%
		\includegraphics[width=0.2\textwidth]{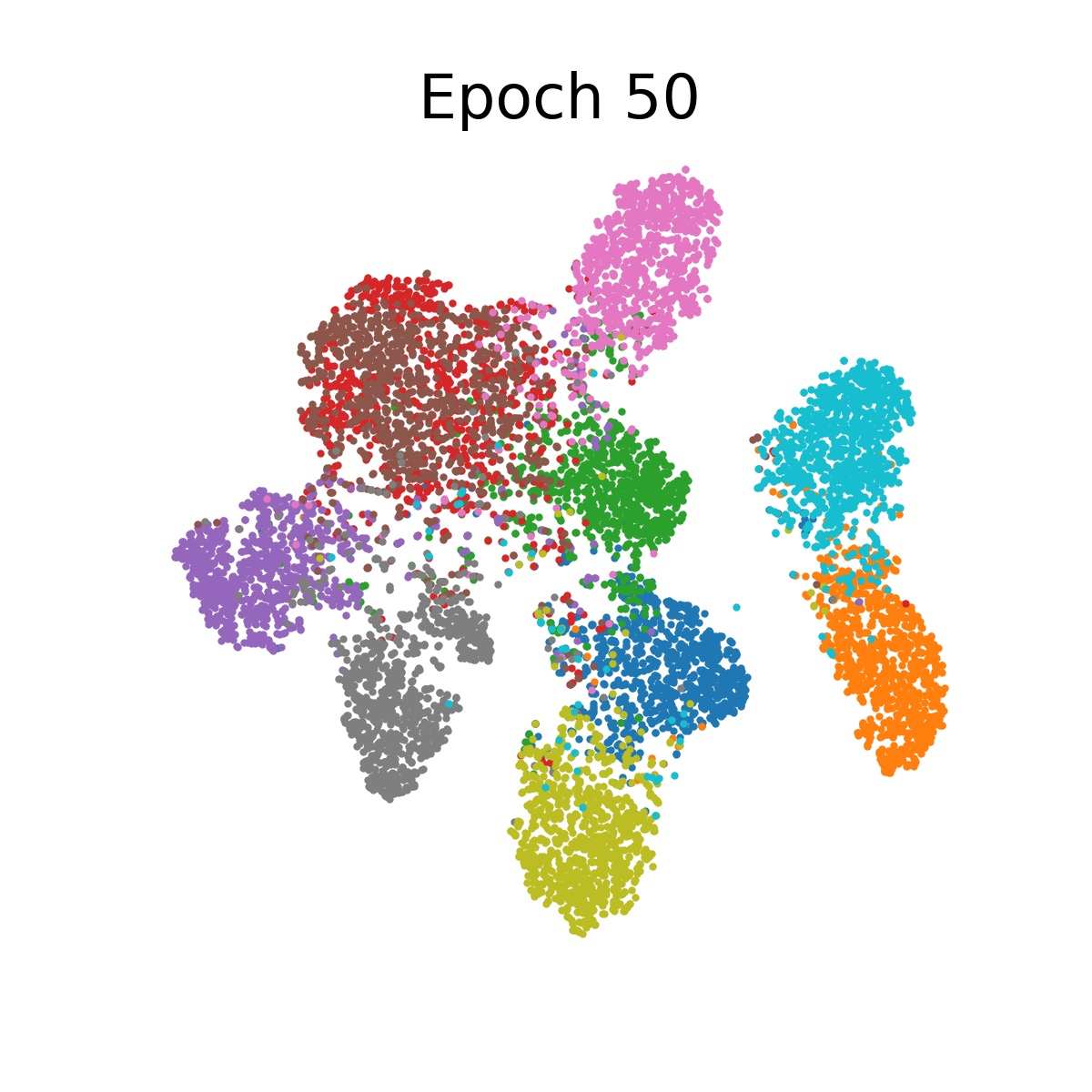}%
		\includegraphics[width=0.2\textwidth]{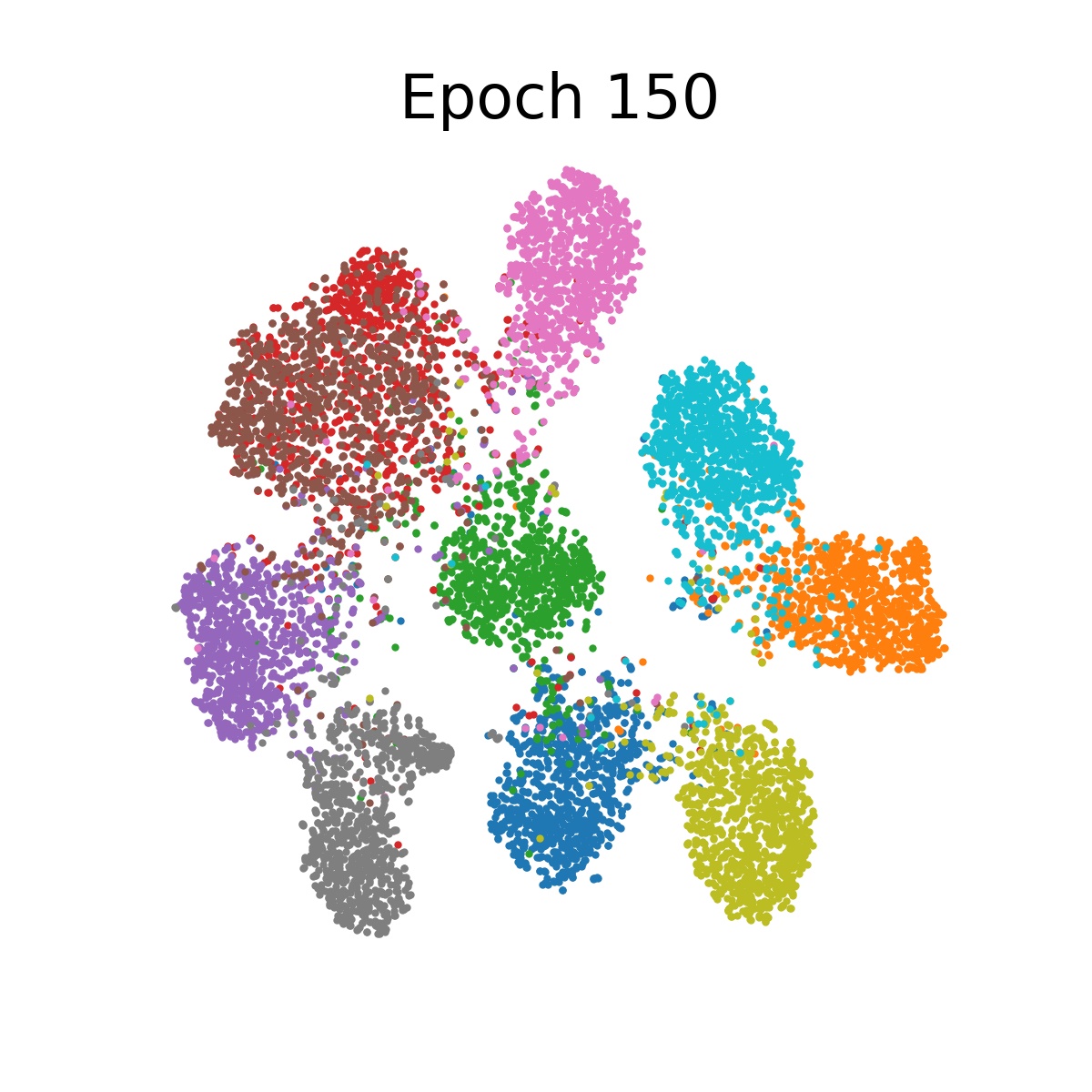}%
		\includegraphics[width=0.2\textwidth]{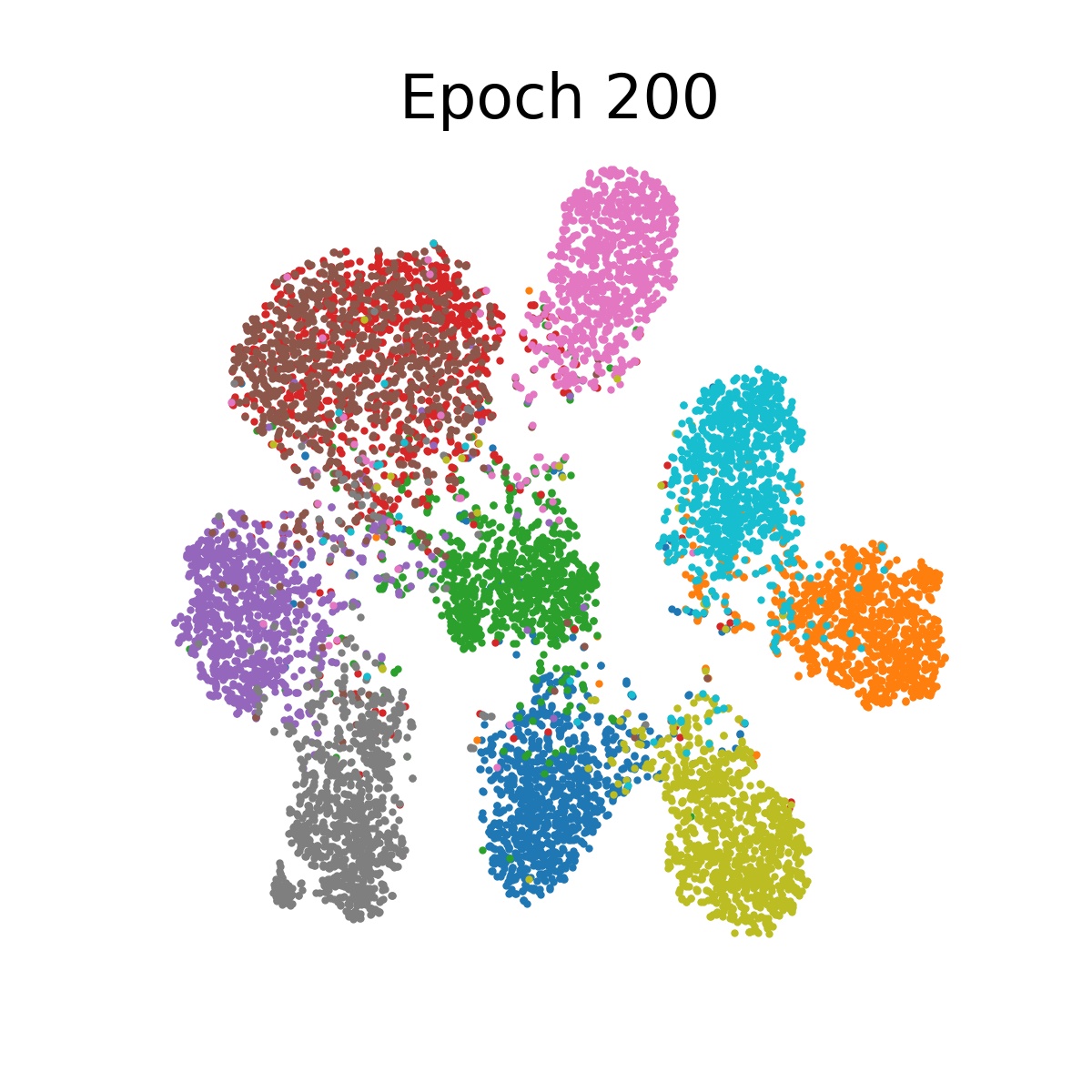}%
		\includegraphics[width=0.2\textwidth]{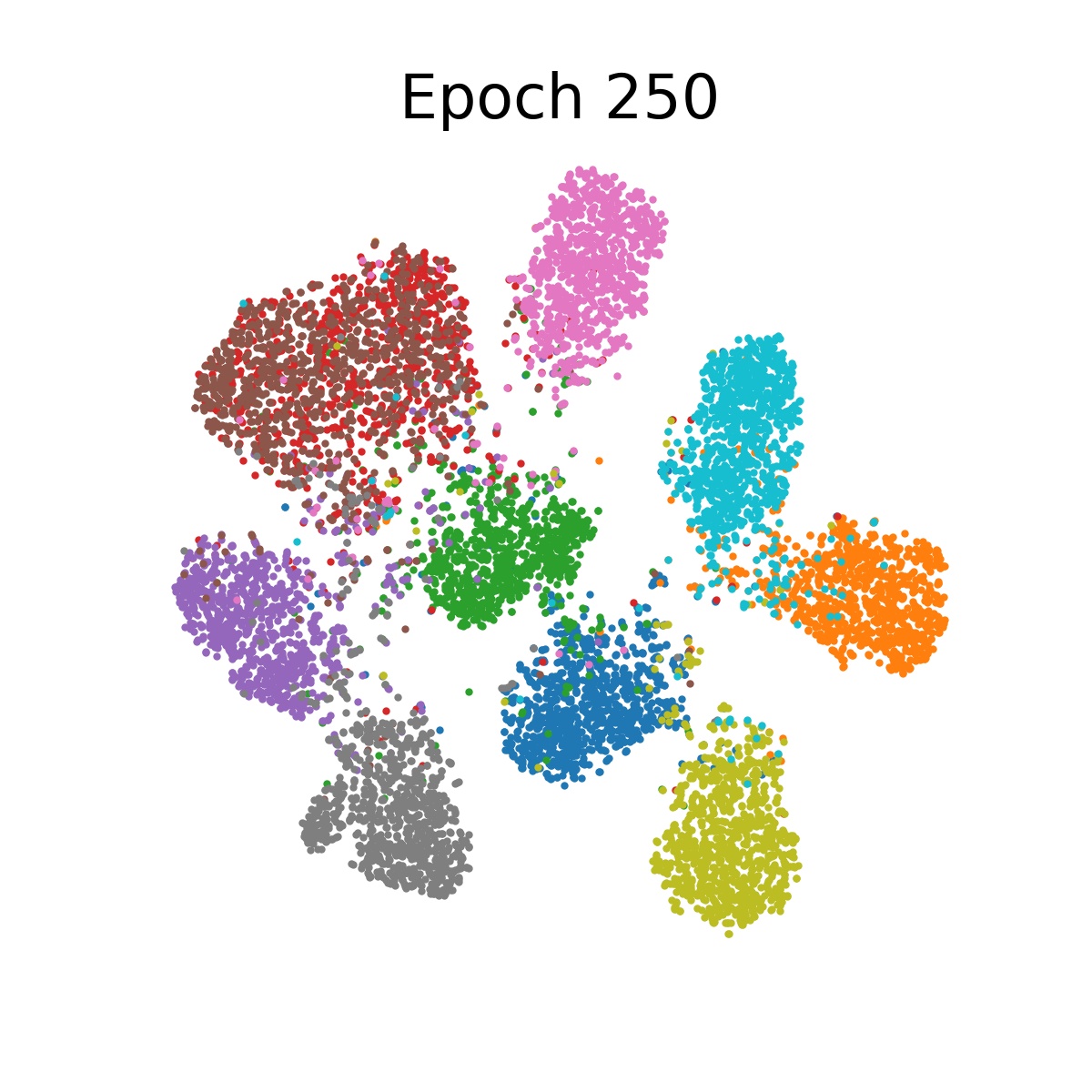}%
	\end{minipage}\vspace{-12pt}
	\begin{minipage}[c]{0.05\columnwidth}\centering\small \rotatebox[origin=c]{90}{-- \footnotesize{Sel-CL} --} \end{minipage}%
	\begin{minipage}[c]{0.95\textwidth}
		\includegraphics[width=0.2\textwidth]{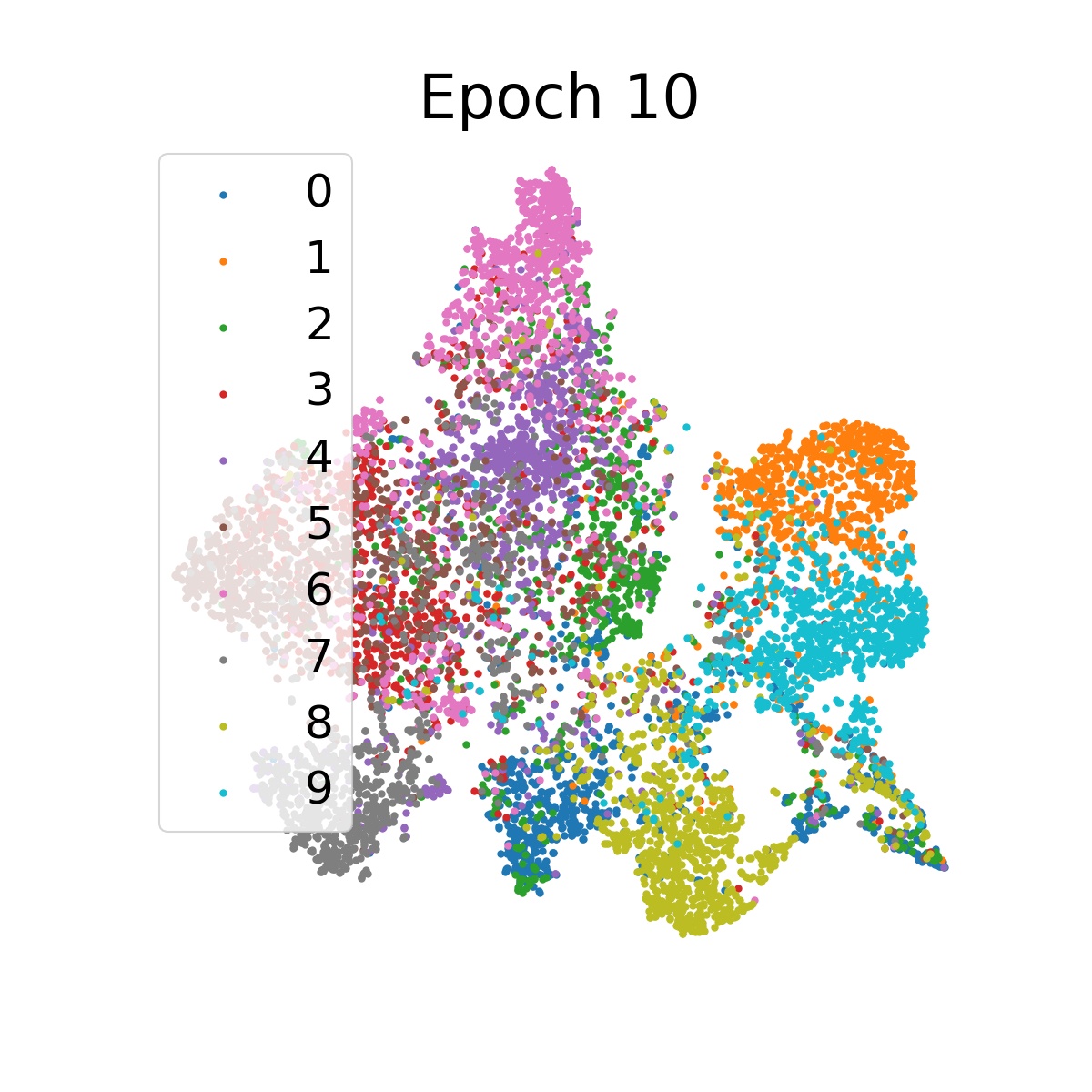}%
		\includegraphics[width=0.2\textwidth]{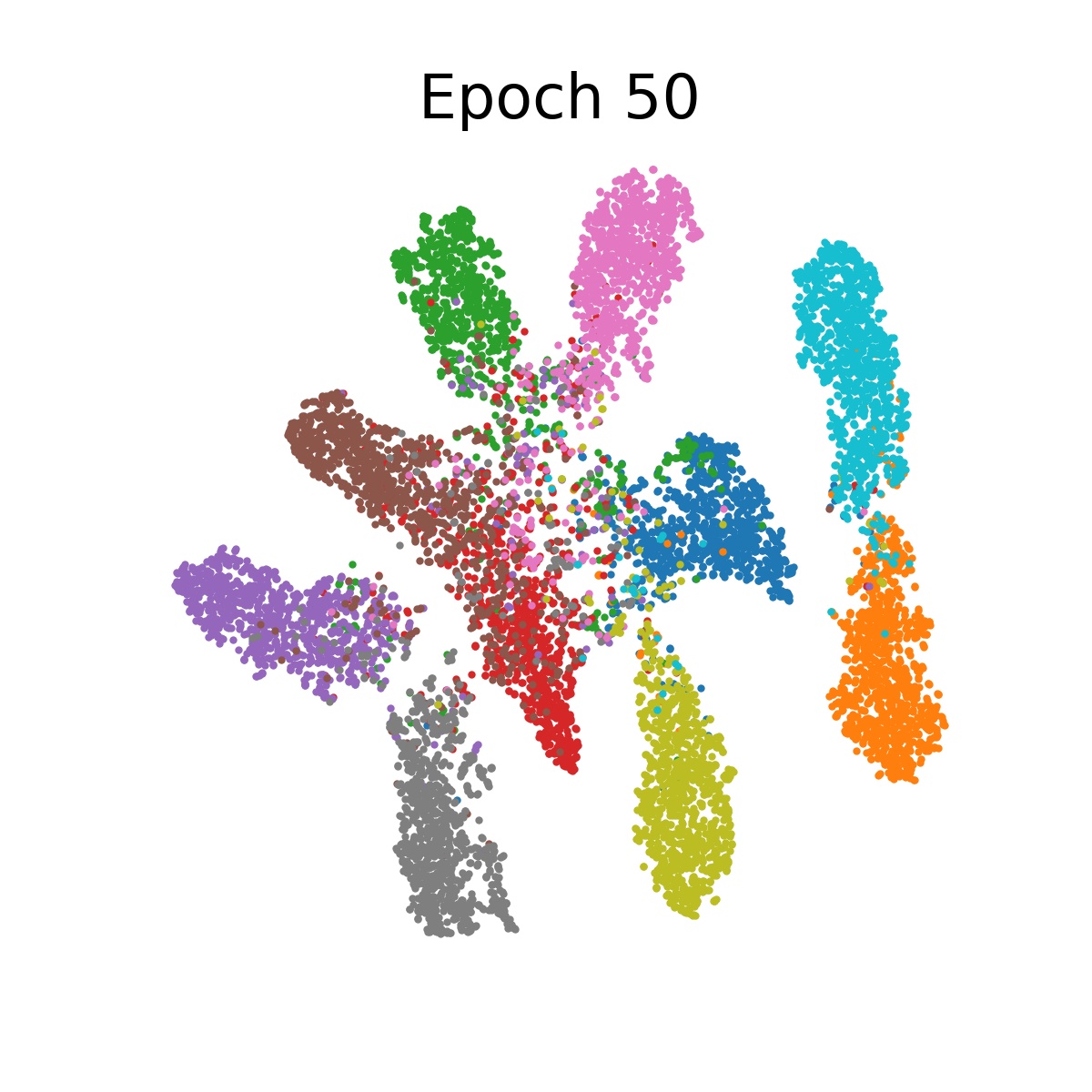}%
		\includegraphics[width=0.2\textwidth]{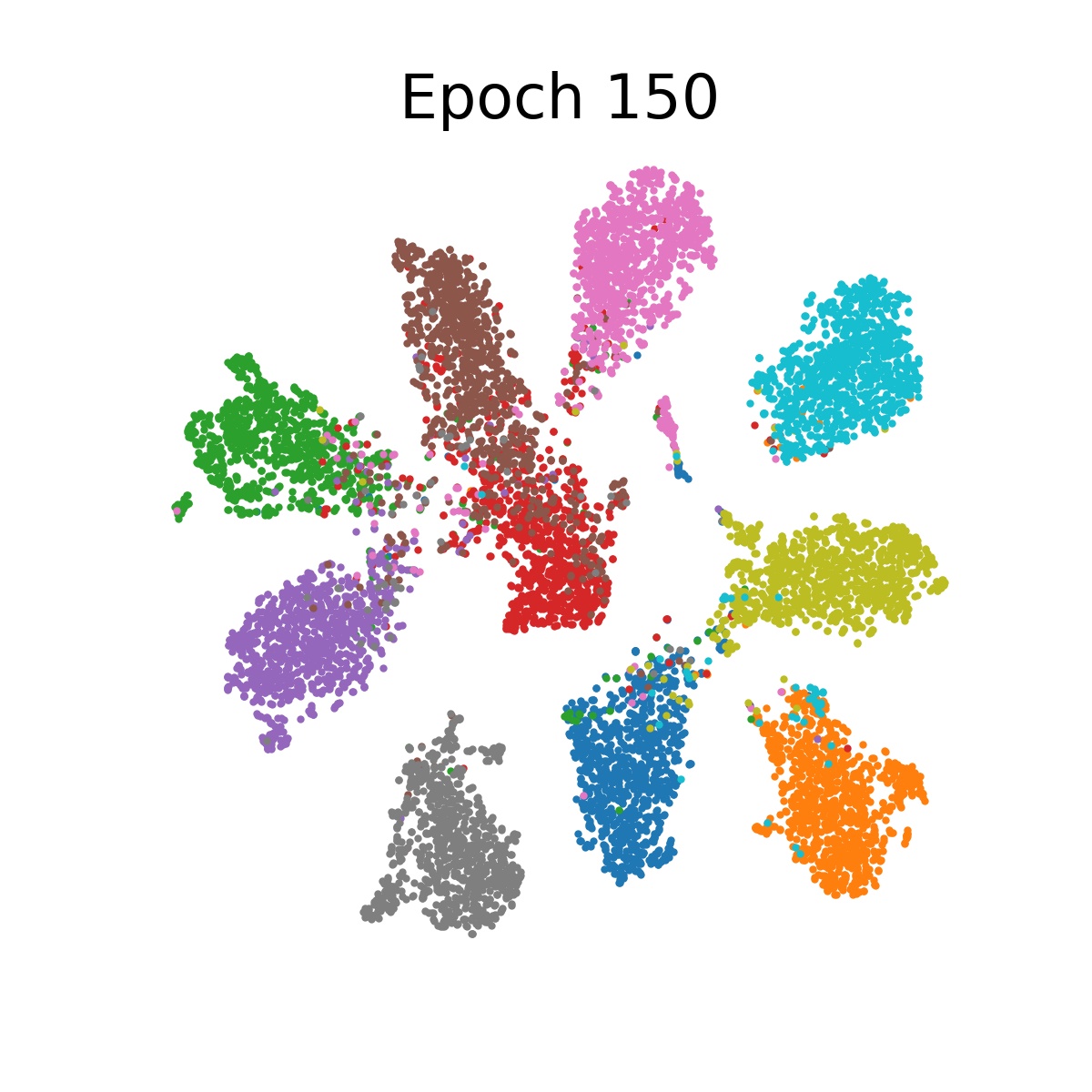}%
		\includegraphics[width=0.2\textwidth]{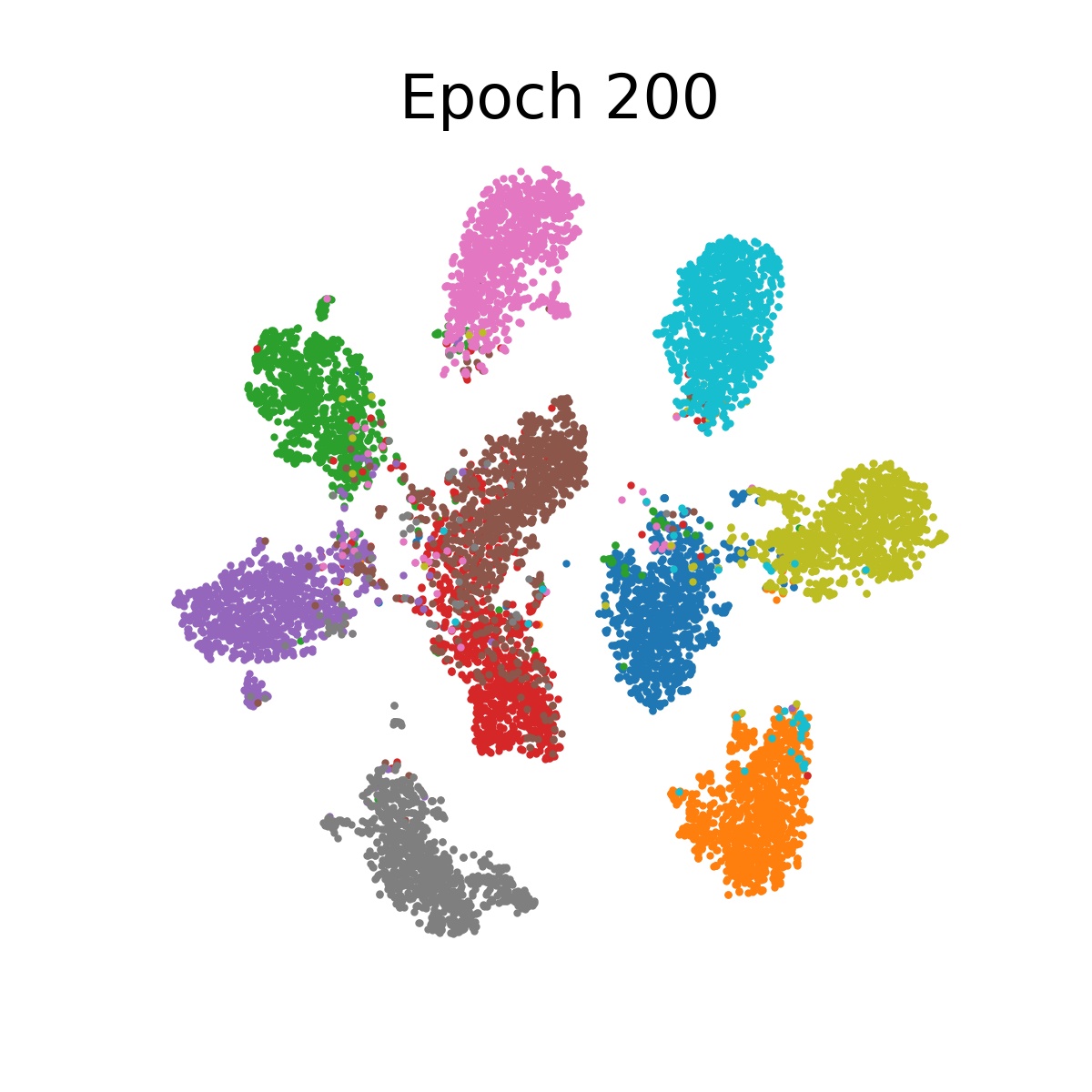}%
		\includegraphics[width=0.2\textwidth]{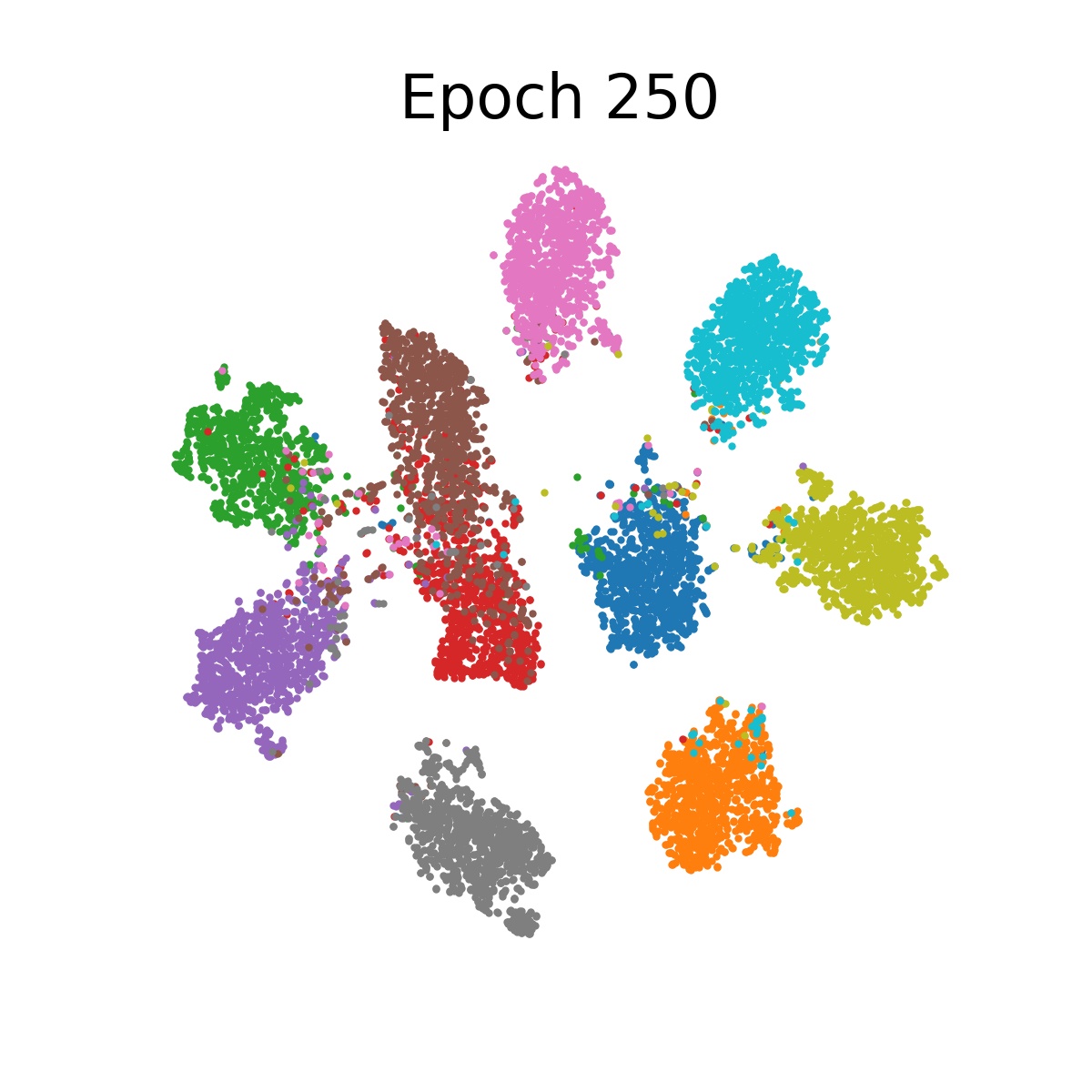}%
	\end{minipage}\vspace{-5pt}
	\caption{t-SNE visualization of representations for CIFAR-10 images. The first two rows denote that the experiments are conducted with 20\% symmetric noise. The last two rows denote that the experiments are conducted with 40\% asymmetric noise.}
	\label{fig:t-sne}
	\vspace{-10pt}
\end{figure*}

\section{Relations with MOIT}
Both MOIT and Sel-CL select confident data in representation learning with noisy labels, but Sel-CL is different from MOIT in two aspects:

(1) The selected targets are different. MOIT perform \textit{point-wise} selection, which aims to select confident examples. While, Sel-CL performs \textit{pair-wise} selection, which aims to select confident pairs. Sel-CL can not only employ the pairs whose class labels are correct, but also use the pairs whose class labels are incorrect.

(2) The roles of selection in representation learning are different. In the pre-training stage, MOIT selects confident examples for performing semi-supervised learning at the classification head. This operation plays a regularization role in representation learning. However, contrastive learning is performed on all noisy pairs.  While, Sel-CL performs contrastive learning on the selected pairs, which improves the robustness of representation learning more directly and effectively. Besides, a series of experiments verify the advantages of Sel-CL compared with MOIT.

\section{Visualization Results}
By using t-SNE visualization, we compare the representations achieved by Cross-Entropy and our method. 
As show in Fig.~\ref{fig:t-sne}, Sel-CL can obtain more robust representations and better combat noisy labels.
%The results are provided in Fig.~\ref{fig:t-sne}, which show that Sel-CL can obtain more robust representations and better combat noisy labels. 

\section{Pseudo-code of the Proposed Sel-CL}
Algorithm~\ref{alg:algorithm1} lists the pseudo-code of Sel-CL. %Besides, the source code of the proposed method is also provided in the file. 

\begin{algorithm}[h]
    \small
	\renewcommand{\algorithmicrequire}{\textbf{Input:}}
	\renewcommand{\algorithmicensure}{\textbf{Output:}}
	\caption{Selective-Supervised Contrastive Learning with Noisy Labels}
	\label{alg:algorithm1}
	\begin{algorithmic}[1]
		\REQUIRE Noisily-labeled dataset $\widetilde{\mathcal{D}}= \{ (\bm{x}_i,\tilde{y}_i )\}_{i=1}^n$, noise detection fractile $\alpha, \beta$, mixup parameter $\alpha_m$, scalar temperature $\tau$, loss weight $\lambda_c,\lambda_s$, warm-up epochs $T_{warm}$ , max epochs $T_{max}$.
		\ENSURE learned deep encoder $f$.
		\FOR {$t=1,2,...,T_{max}$}
		\IF{$t\leq T_{warm}$}{
			\STATE Train the deep encoder $f$ with Uns-CL or Sup-CL.}
		\ELSE
		\STATE Selecting confident examples $\mathcal{T}$ by measuring the agreement between learned representations and given noisy labels.
		
		\STATE Selecting confident pairs $\mathcal{G}$ by exploiting representation similarity distribution in $\mathcal{T}$.
		\STATE Train the deep encoder $f$ by performing supervised contrastive learning on $\mathcal{G}$ and classfication learning on $\mathcal{T}$ with Mixup technique.
		\ENDIF
		\ENDFOR
		\STATE \textbf{return} deep encoder $f$.
	\end{algorithmic}
\end{algorithm}

%%%%%%%%% REFERENCES
%{\small
%\bibliographystyle{ieee_fullname}
%\bibliography{bibSelCon}
%}